\documentclass{article}

\usepackage[preprint]{neurips_2025}

\usepackage[utf8]{inputenc} 
\usepackage[T1]{fontenc}    
\usepackage{hyperref}       
\usepackage{url}            
\usepackage{booktabs}       
\usepackage{amsfonts}       
\usepackage{nicefrac}       
\usepackage{microtype}      
\usepackage{xcolor}         
\usepackage{graphicx}
\usepackage{colortbl}
\usepackage{multirow} 
\usepackage{subcaption}
\usepackage{wrapfig}
\usepackage{marvosym}

\title{TinyLLaVA-Video: Towards Smaller LMMs for Video Understanding with Group Resampler}

\author{\hspace{-5pt}Xingjian Zhang$^{1}$\quad Xi Weng$^{1}$\quad Yihao Yue$^{1}$\quad Zhaoxin Fan$^{1,2}$\quad Wenjun Wu$^{1,2,3}$\quad Lei Huang$^{1,2,3,~\textrm{\Letter}}$
\\
\\
{\small$^{1}$SKLCCSE, Institute of Artificial Intelligence, Beihang University, Beijing, China}\\
{\small $^{2}$Beijing Advanced Innovation Center for Future Blockchain and Privacy Computing, Beihang University}\\
{\small $^{3}$Hangzhou International Innovation Institute, Beihang University, Hangzhou, China}\\
\normalsize\texttt{\{huangleiai\}@buaa.edu.cn}
}

\begin{document}
\maketitle
\begin{abstract}

Video behavior recognition and scene understanding are fundamental tasks in multimodal intelligence, serving as critical building blocks for numerous real-world applications. Through Large Multimodal Models (LMMs) have achieved remarkable progress in video understanding, most existing open-source models rely on over 7B parameters and require large-scale datasets for training, making them resource-intensive and inaccessible to many researchers. Furthermore, lightweight models face persistent challenges in effectively processing long visual sequences and temporal understanding. In this work, we introduce TinyLLaVA-Video, a lightweight yet powerful video understanding model with approximately 3.6B parameters. The cornerstone of our design is the video-level group resampler, a novel mechanism that significantly reduces and controls the number of visual tokens at the video level. Unlike traditional image-level resampler, our approach effectively mitigates redundancy while enhancing temporal comprehension, leading to improved performance on video-based tasks. In addition, TinyLLaVA-Video demonstrates exceptional efficiency, requiring only one day of training on 8 A100-40G GPUs. It surpasses several existing 7B-parameter models on multiple benchmarks. We believe this work provides a valuable foundation for future research on lightweight video understanding models. The code and weights is available at \url{https://github.com/ZhangXJ199/TinyLLaVA-Video}.

\end{abstract}
\vspace{-0.1in}
\section{Introduction}
\label{sec:intro}
\vspace{-0.05in}

\begin{wrapfigure}{r}{0.61\textwidth}
  \vspace{-0.2in}
  \centering
  \includegraphics[width=\linewidth]{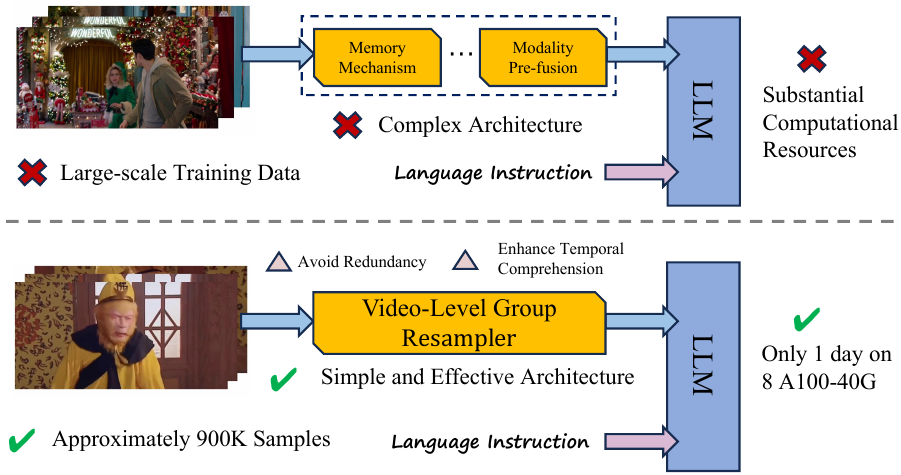}
   \caption{Compared to other pipelines, our method is simple and effective, reducing the required computational resources.}
   \label{fig:advantage}
   \vspace{-0.05in}
\end{wrapfigure}

Video understanding is a cornerstone of multimodal intelligence, enabling critical applications such as action recognition, dynamic scene understanding, and video-text interaction. These capabilities are essential for advancing fields like autonomous driving, security surveillance, and human-computer interaction, where the ability to comprehend both spatial and temporal information is paramount. Despite its importance, video understanding remains one of the most challenging problems in computer vision due to the inherent complexity of processing long visual sequences and capturing temporal dynamics.

Existing approaches to video understanding have made progress but face significant limitations. Due to the input-length limitations inherent in language models, prior methods typically restrict the number of extracted frames\cite{liu2024llava} or introduce additional video encoders\cite{lin2023video}. However, these solutions either severely limit the acquisition of visual information or result in more complex model architectures and an increased number of parameters. Inspired by Flamingo~\cite{alayrac2022flamingo}, many methods focus on resampling individual video frames~\cite{ye2024mplug, bai2023qwen, zhang2025llava}. Although this approach effectively reduces token counts, it overlooks the temporal relationships between video frames, which is critical for recognizing complex actions and understanding dynamic scenes. In addition, most of these models are based on large-scale datasets and contain more than 7B parameters, placing them out of reach for researchers with limited computational resources.


\begin{wrapfigure}{r}{0.4\textwidth}
  \vspace{-0.2in}
  \centering
  \includegraphics[width=\linewidth]{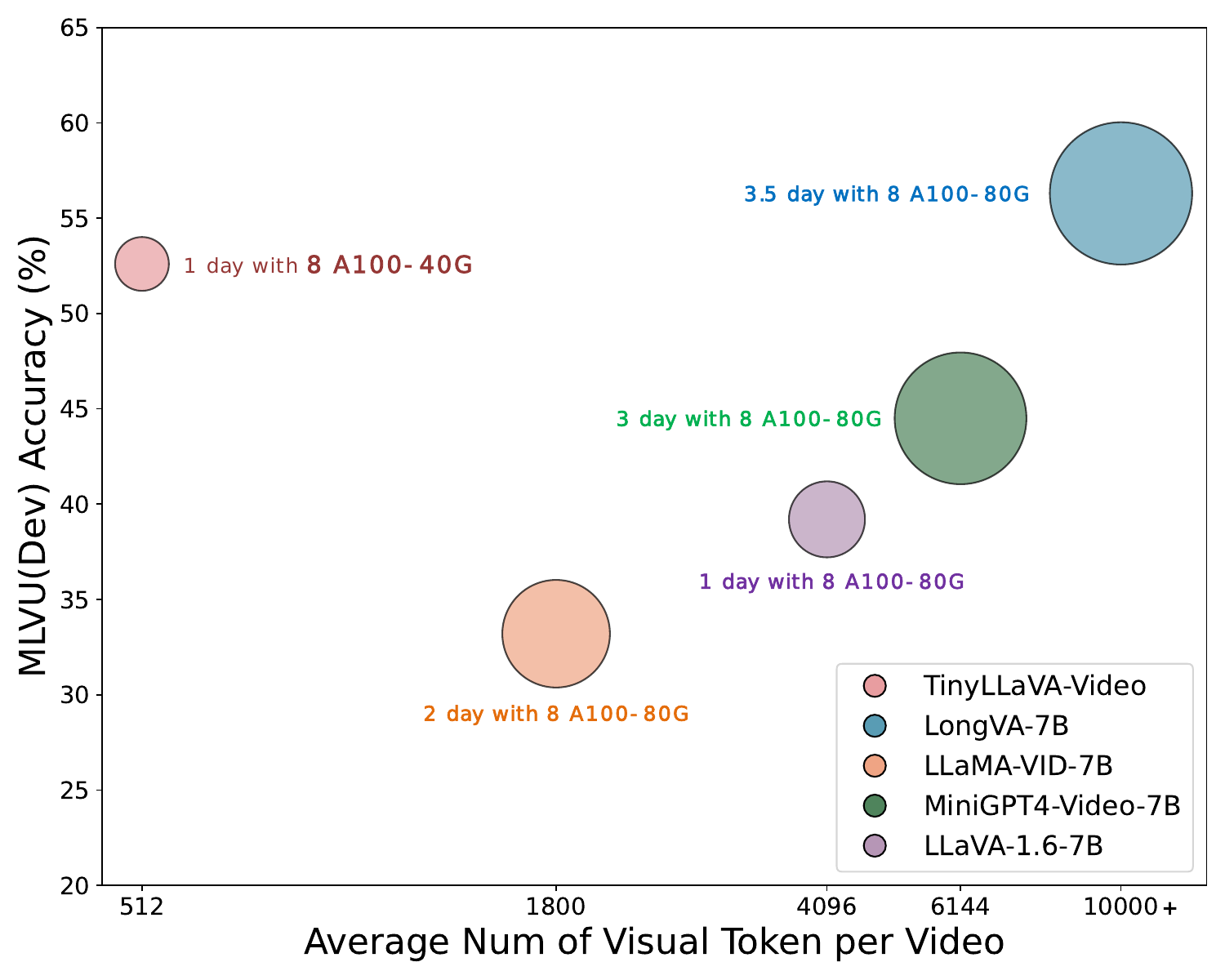}
   \caption{The model performance on MLVU (Dev) versus the total number of visual tokens used per video, with bubble size indicating the training resources required for each model.}
   \label{fig:video_model}
   \vspace{-0.15in}
\end{wrapfigure}

In light of these challenges, we aim to develop a simple yet effective model that significantly reduces the number of visual tokens at the video sequence level while enhances temporal comprehension across frames. Additionally, our goal is to minimize the reliance on computational resources during training, making video understanding more accessible to researchers with limited resources.

To this end, we propose a novel lightweight model termed TinyLLaVA-Video. At the core of our design is the video-level group resampler, a novel mechanism that groups learnable queries with the entire visual sequence. This approach significantly reduces the number of visual tokens at the video sequence level while ensuring efficient query learning. Unlike commonly used image-level resampler, which process frames independently, the video-level group resampler enhances interactions across adjacent frames, capturing critical temporal dependencies. This design is simple yet effective, avoiding complex model architectures. In addition, we limit our training to only 900K samples to reduce resource requirements while ensuring performance remains competitive (as shown in Fig.~\ref{fig:advantage}).

With these novel designs, TinyLLaVA-Video achieves performance comparable to several existing 7B-parameter models, while requiring just one day of training on 8 A100-40G GPUs (as shown in Fig.~\ref{fig:video_model}). We believe this work will not only empower researchers with limited computational resources, but also serve as a foundational reference for future studies on lightweight video understanding models. Our contributions can be summarized as:
\begin{itemize}
    \item We propose TinyLLaVA-Video, a small-scale model with 3.6B parameters, which requires only one day of training on 8 A100-40G GPUs, significantly lowering computational resource barriers.

    \item We point out the redundancy in the resampler and propose a novel video-level group resampler, which significantly reduces the number of visual tokens while preserving the efficiency of query learning and enhances the understanding of temporal information.

    \item  We conduct extensive experiments to validate the superiority of TinyLLaVA-Video, demonstrating that it outperforms several existing 7B models on multiple benchmarks.
\end{itemize}


\vspace{-0.1in}
\section{Related Work}
\label{sec:related}

\vspace{-0.05in}
\paragraph{Small-scale Language Models.} While the advancements in large language models in recent years are well acknowledged, small-scale language models have garnered increasing attention from researchers with limited computational resources due to the high costs associated with deploying large models. Currently, a number of high-performing small-scale language models, such as Qwen2.5~\cite{hui2024qwen2}, Phi-2~\cite{javaheripi2023phi}, Gemma~\cite{team2024gemma}, and Tinyllama~\cite{zhang2024tinyllama}, are widely utilized. These models deliver robust performance despite computational constraints, thereby fostering the continued progress of small-scale multimodal models.

\vspace{-0.1in}
\paragraph{Small-scale Multimodal Models.} Since the introduction of fine-tuning LMMs with visual instruction tuning data by LLaVA~\cite{liu2024visual}, researchers have observed that reducing the size of language models within LLaVA can still yield satisfactory performance. This has led to the emergence of small-scale multimodal models such as TinyLLaVA~\cite{zhou2024tinyllava}, Bunny~\cite{he2024efficient}, and Mini-Gemini~\cite{li2024mini}. These models have achieved competitive results on benchmarks such as GQA~\cite{hudson2019gqa} and MMMU~\cite{yue2024mmmu} for image-text understanding. However, most fully open-source multimodal models with fewer than 4B parameters remain limited to image understanding and struggle to comprehend video sequences. 

\begin{figure}[t]
  \centering
  \includegraphics[width=\textwidth]{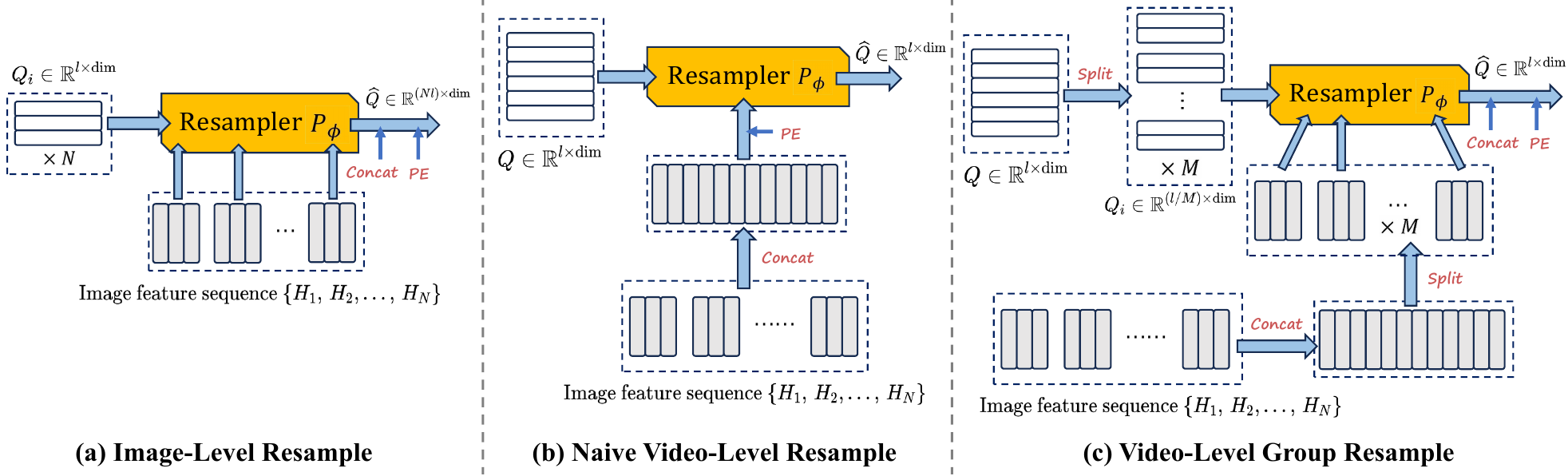}
   \caption{Comparison of different resampler. The total number of queries obtained by the image-level resampler is related to the number of sampled frames. In contrast, both video-level methods reduce the number of visual tokens at the entire video sequence level, making the total count limited and more controllable. PE and Concat represent Position Encoding and Concatenate, respectively.}
   \label{fig:resample}
   \vspace{-0.15in}
\end{figure}

\vspace{-0.1in}
\paragraph{Video Understanding Models.} The challenge of processing video sequences arises from the context-length limitations of language models. LLaVA-NEXT~\cite{liu2024llava} allocates the number of frames within the token limit through linear adjustments, but this severely restricts the amount of visual information that can be obtained. Video-LLaVA~\cite{lin2023video} designs multimodal encoders to align video data with the textual feature space, and MovieChat~\cite{song2024moviechat} incorporates a memory mechanism for storage, both of which introduce additional architectural complexity. Video-ChatGPT~\cite{maaz2023video} and VideoLLaMA~\cite{zhang2023video} adopt pooling and convolution techniques to reduce the number of visual tokens. Inspired by Flamingo~\cite{alayrac2022flamingo}, LLaMA-VID~\cite{li2025llama} and LLAVA-MINI~\cite{zhang2025llava} significantly reduce the number of visual tokens per image, and we categorize their methods as image-level resampling. However, LLAVA-MINI also incorporates a pre-fusion module to integrate more visual information. For small-scale models, we aim to achieve effective video understanding with a lightweight architecture that relies on a controlled and limited number of visual tokens.

\vspace{-0.1in}
\section{TinyLLaVA-Video Design Choice}

\begin{wrapfigure}{r}{0.58\textwidth}
\vspace{-0.2in}
  \centering
  \includegraphics[width=\linewidth]{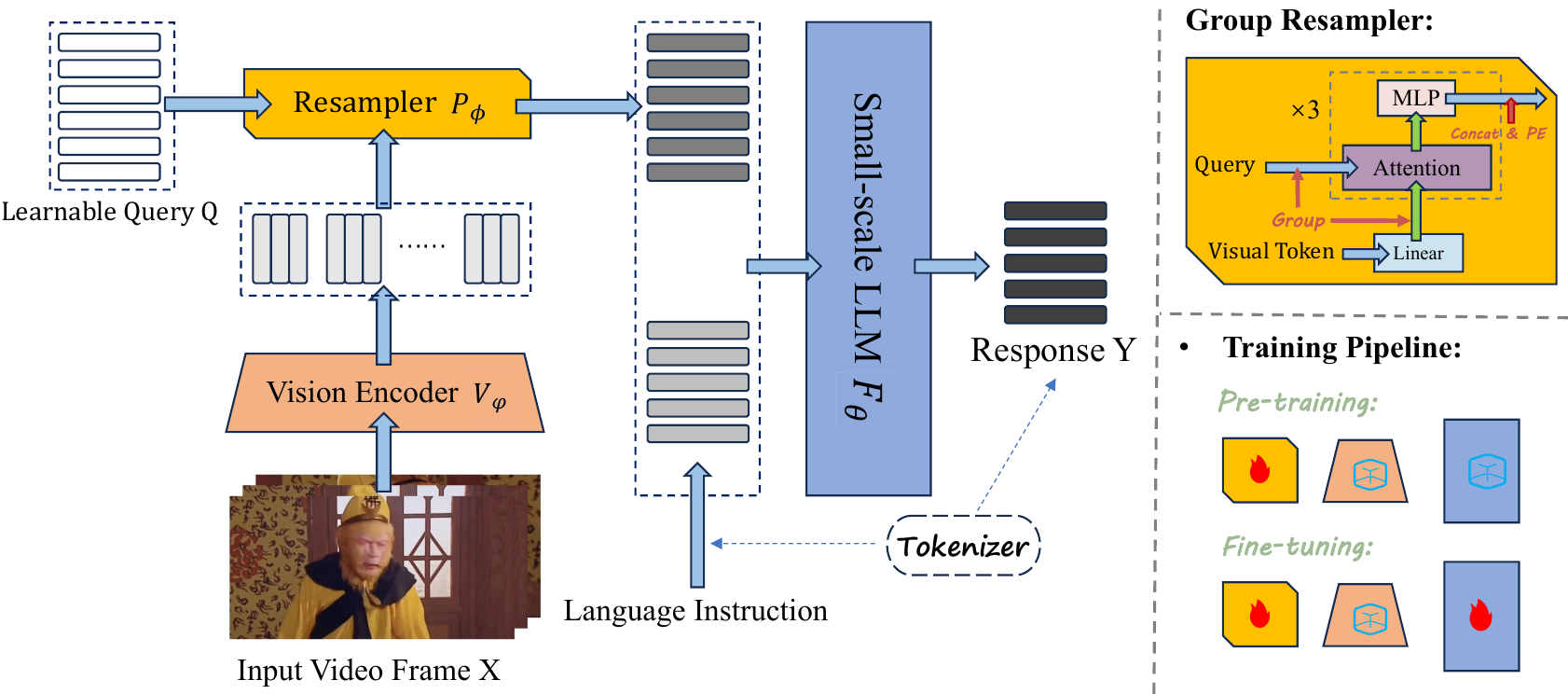}
   \caption{The structure and pipeline of TinyLLaVA-Video.}
   \label{fig:architecture}
   \vspace{-0.1in}
\end{wrapfigure}

This section will present the design choices of TinyLLaVA-Video, introduce the model architecture, and analyze three different resampling methods, followed by an overview of the training pipeline and training data, as shown in Fig.~\ref{fig:architecture}. Without requiring additional modules, the model resamples the entire sequence of visual tokens in a simple manner, enabling effective temporal comprehension using a controllable and limited number of visual tokens.

\vspace{-0.1in}
\subsection{Model Architecture}

For video sequences, TinyLLaVA-Video allows customization of the number of extracted frames, supporting both fps sampling and uniform frame sampling with the extracted sequence \(\{{X}_1, {X}_2, \dots, {X}_N \}\) as input. In fps sampling, frames are extracted from the video based on the duration and frame rate, while in uniform frame sampling, frames are selected at uniform intervals from the video based on the desired number of samples.

\vspace{-0.1in}
\paragraph{Vision Encoder \( V_\varphi \).} Vision encoder \( V_\varphi \) extracts features for each image: \({H}_i = {V}_\varphi({X}_i)\), producing a visual feature sequence \(\{{H}_1, {H}_2, \dots, {H}_N \}\). The visual encoder can be CLIP~\cite{radford2021learning}, SigLIP~\cite{zhai2023sigmoid} or Dinov2~\cite{oquab2023dinov2}. 

\begin{wrapfigure}{r}{0.43\textwidth}
  \centering
  \includegraphics[width=\linewidth]{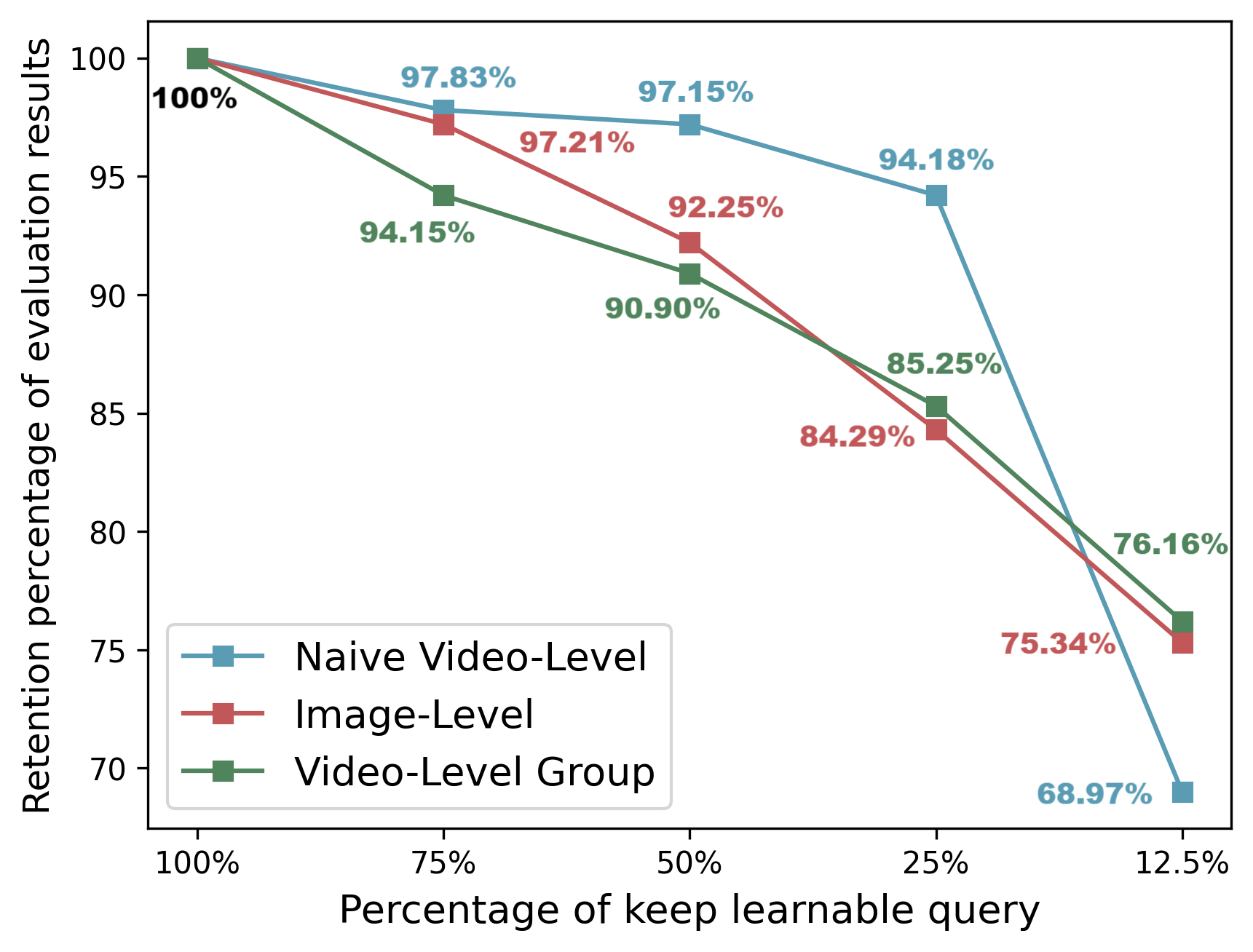}
   \caption{Performance degradation on Video-MME-Short of different resampler when zeroing out partial queries. The results show that the query learning efficiency in naive video-level method is relatively low, resulting in noticeable redundancy.}
   \label{fig:line_three}
   \vspace{-0.4in}
\end{wrapfigure}

\vspace{-0.1in}
\paragraph{Connector (Resampler) \( P_\phi \).} In the connector module, we set learnable queries Q to resample the visual features. The number of queries is significantly reduced compared to the original visual tokens. Subsequently, these queries will be fed into the language model. 

\vspace{-0.1in}
\paragraph{Small-scale LLM \( F_\theta \).} The input to the small-scale language model consists of text tokens obtained through the tokenizer and embedding module, along with the queries Q, both having the same dimensions. After processing through the language model, the output is the response Y. 

\vspace{-0.05in}
\subsection{Analysis of Resampling Methods}
\vspace{-0.05in}
\label{analysis}

\begin{figure}[t]
  \centering
  \begin{subfigure}[b]{0.32\textwidth}
    \includegraphics[width=\linewidth]{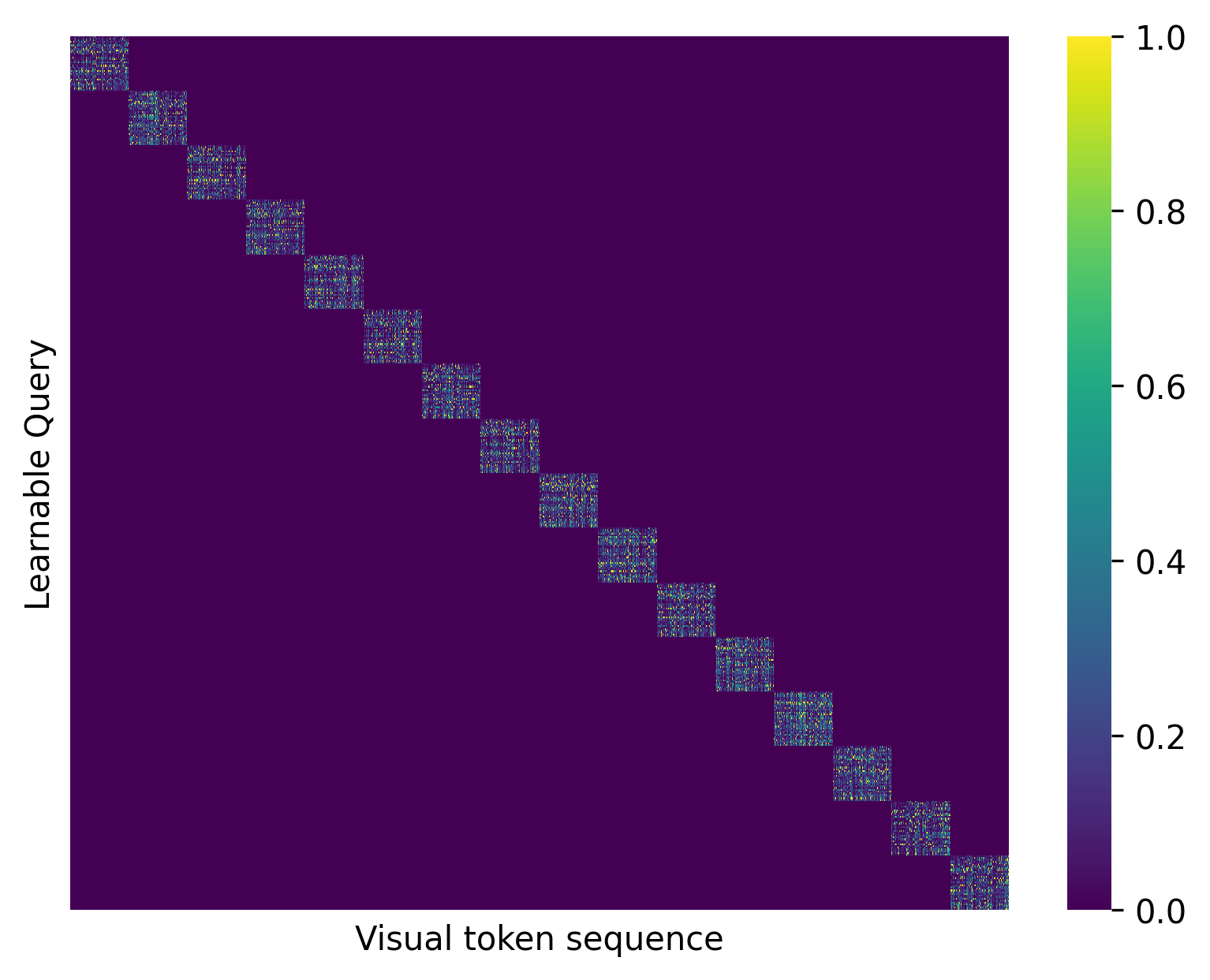}
    \caption{\small Image-Level Resample.}
    \label{fig:heatmap_im}
  \end{subfigure}
  \hfill
  \begin{subfigure}[b]{0.32\textwidth}
    \includegraphics[width=\linewidth]{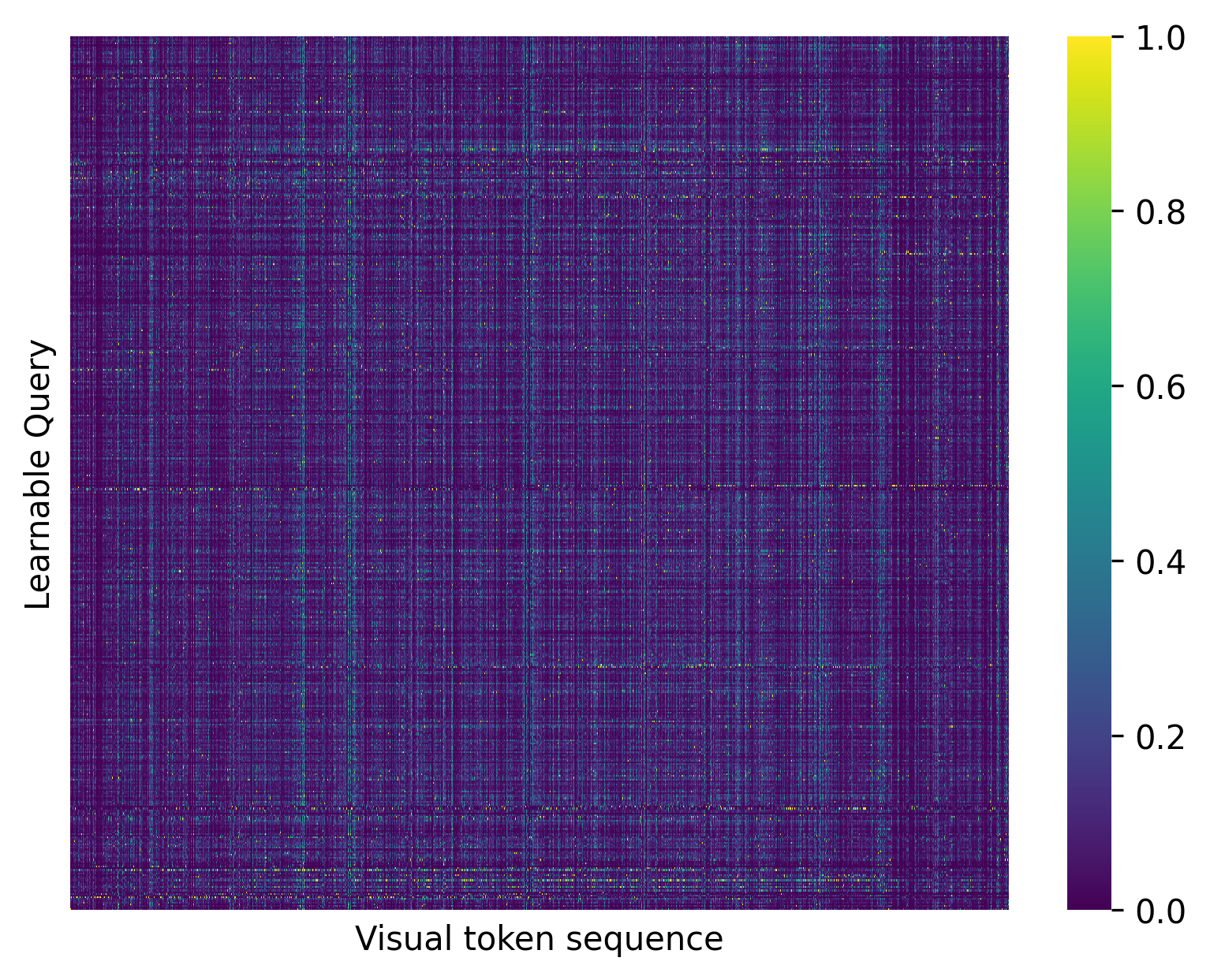}
    \caption{\small Naive Video-Level Resample.}
    \label{fig:heatmap_naive}
  \end{subfigure}
  \hfill
  \begin{subfigure}[b]{0.32\textwidth}
    \includegraphics[width=\linewidth]{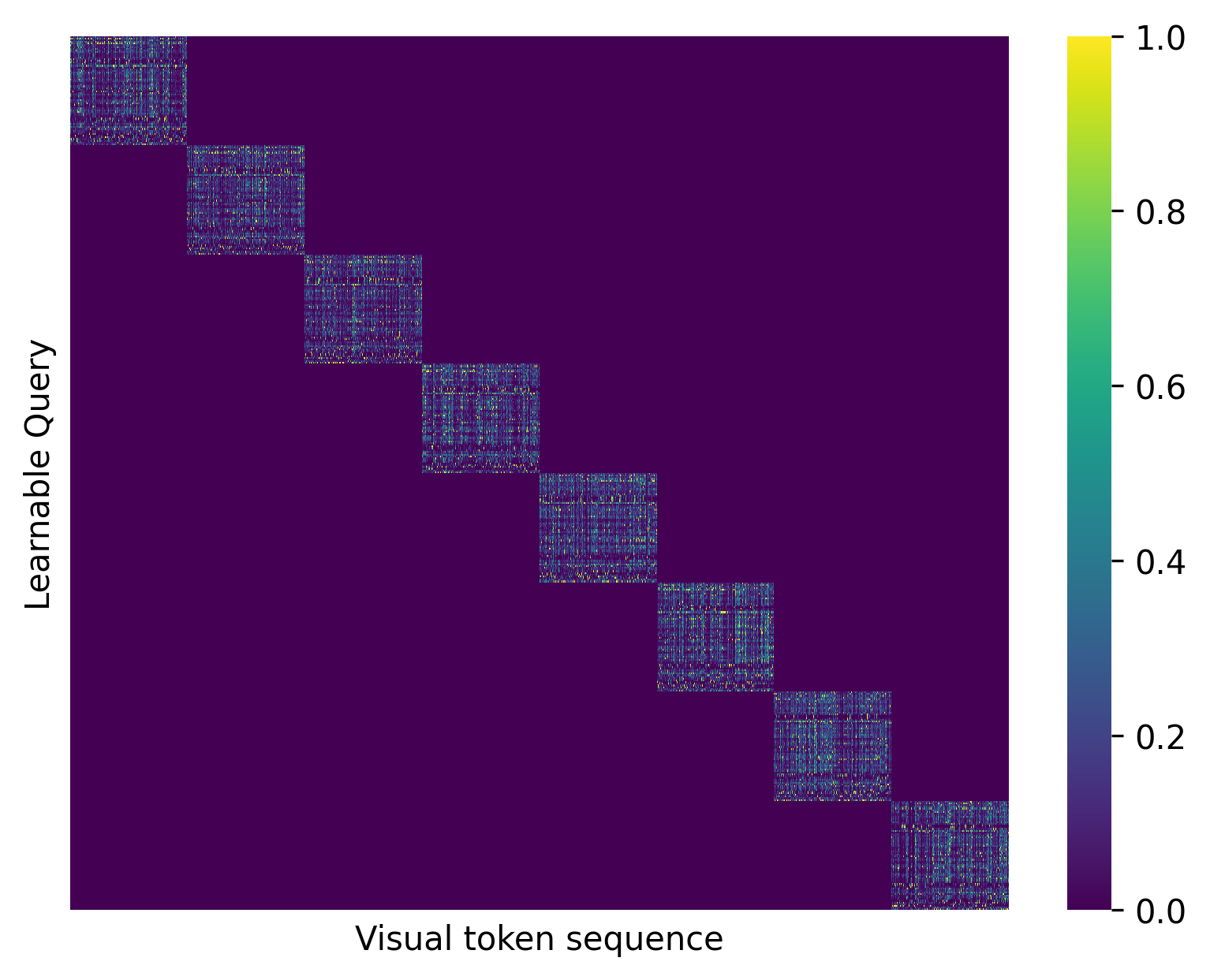}
    \caption{\small Video-Level Group Resample.}
    \label{fig:heatmap_group}
  \end{subfigure}
  \caption{Attention visualization between the learnable queries and the visual token sequence in the last layer. In Naive Video-Level method, the queries pay adequate attention to the entire input sequence. In contrast, the other two methods are only responsible for the corresponding parts, resulting in higher learning efficiency.}
  \label{fig:heatmap}
  \vspace{-0.15in}
\end{figure}

Next, we will introduce and analyze three different resampling methods, whose structures are shown in Fig.~\ref{fig:resample}. Existing methods~\cite{bai2023qwen, zhang2025llava} typically perform resampling on each video frame, which is an image-level resampling approach as follows.

\begin{figure*}[t]
  \centering
  \includegraphics[width=\textwidth]{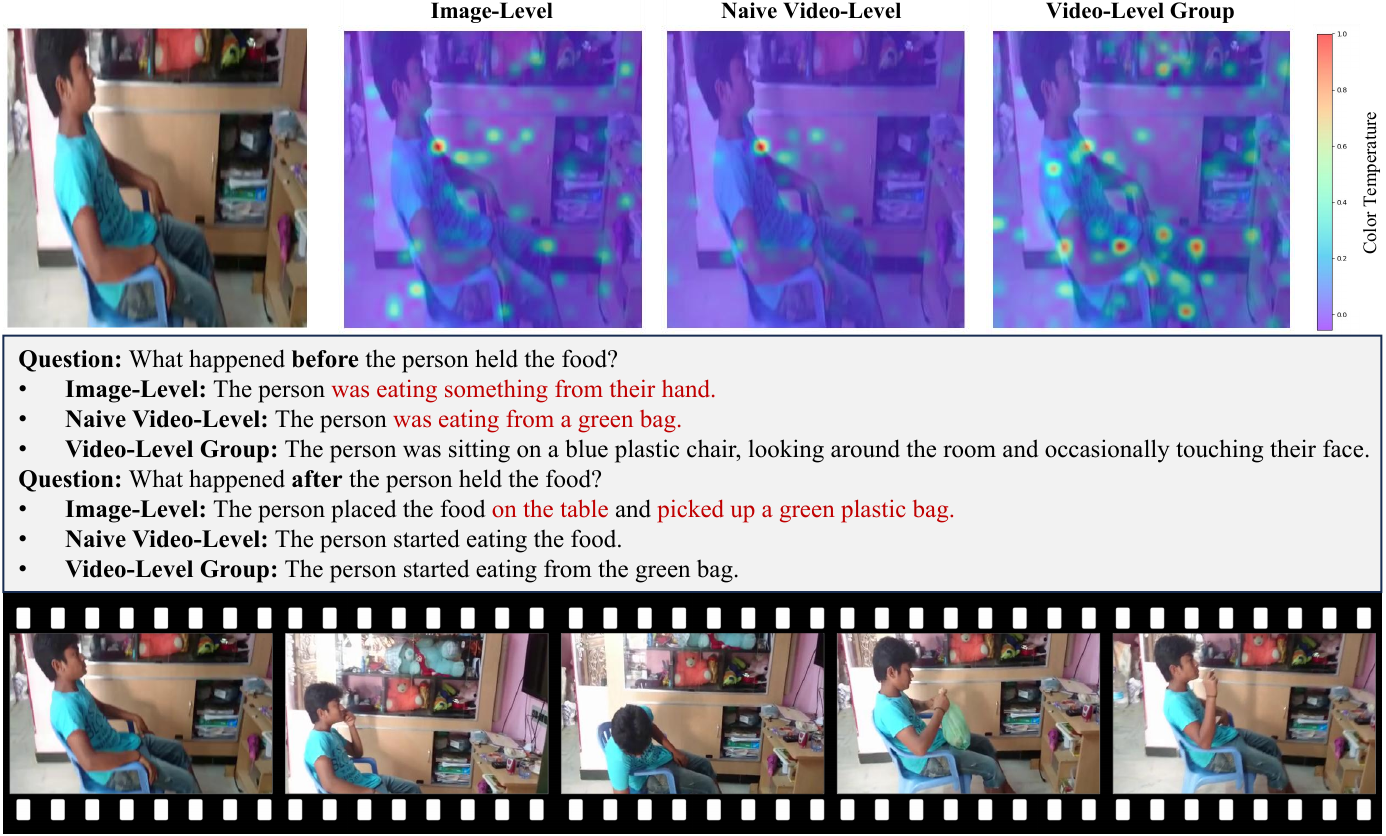}
   \caption{Attention visualization and temporal comprehension capability comparison of three resampler. Naive video-level resampler has sparser attention and weaker image information capture.  Since the queries in image-level resampler only focus on the corresponding image, its temporal understanding capability is weaker. Instead, video-level group resampler has strong image information capturing ability and can also understand the temporal relationships in the video. The case is from the action sequence task in MVBench. The content marked in red indicates errors.}
   \label{fig:attncase}
   \vspace{-0.15in}
\end{figure*}

\vspace{-0.05in}
\paragraph{Image-Level Resample.} For each image feature \( {H}_i \) in the visual sequence, it is directly fed into the resampler, where a fixed number of learnable queries are used to resample it:
\begin{equation}
\hat{H}_i = \mathit{Linear}({H}_i) .
\label{eq:1}
\end{equation}
\begin{equation}
\hat{Q}_i = \mathit{CrossAttention}(\{q={Q}_i,\,k,v=\hat{H}_i \}) .
\label{eq:2}
\end{equation}
Then, the resampled queries responsible for different images  are concatenated:
\begin{equation}
\hat{Q} = \mathit{Concat}(\{\hat{Q}_1, \hat{Q}_2, \dots, \hat{Q}_N \}) .
\label{eq:3}
\end{equation}
This approach ensures that the queries are responsible only for learning the corresponding image feature. We visualize the attention between the learnable queries and the visual token sequence. As shown in Fig.~\ref{fig:heatmap_im}, high attention values are only concentrated on the corresponding visual token sequence. As a result, when the number of sampled video frames increases, the total number of queries also increases accordingly. Meanwhile, we observe that this method makes the model prone to ignore temporal relationships between video frames, as illustrated in the case in Fig.~\ref{fig:attncase}.

To perform video understanding using a controllable and limited number of visual tokens, while mitigating the impact of the number of sampled frames on the total token count and enhancing temporal understanding, we attempt to resample all extracted video frames as a whole. First, we explore the naive video-level resampling method as follows.

\vspace{-0.05in}
\paragraph{Naive Video-Level Resample.} Before passing the visual feature sequence \(\{{H}_1, {H}_2, \dots, {H}_N \}\) into the resampler \( P_\phi \), it is concatenated into a long tensor:
\begin{equation}
\hat{H} = \mathit{Linear}(\mathit{Concat}(\{{H}_1, {H}_2, \dots, {H}_N \})) .
\label{eq:4}
\end{equation}
Then, learnable queries Q are initialized to learn the overall features of the visual feature sequence, with their number preset as a fixed value:
\begin{equation}
\hat{Q} = \mathit{CrossAttention}(\{q=Q,\,k,v=\hat{H} \}) .
\label{eq:5}
\end{equation}
\begin{wrapfigure}{r}{0.43\textwidth}
  \vspace{-0.2in}
  \centering
  \includegraphics[width=\linewidth]{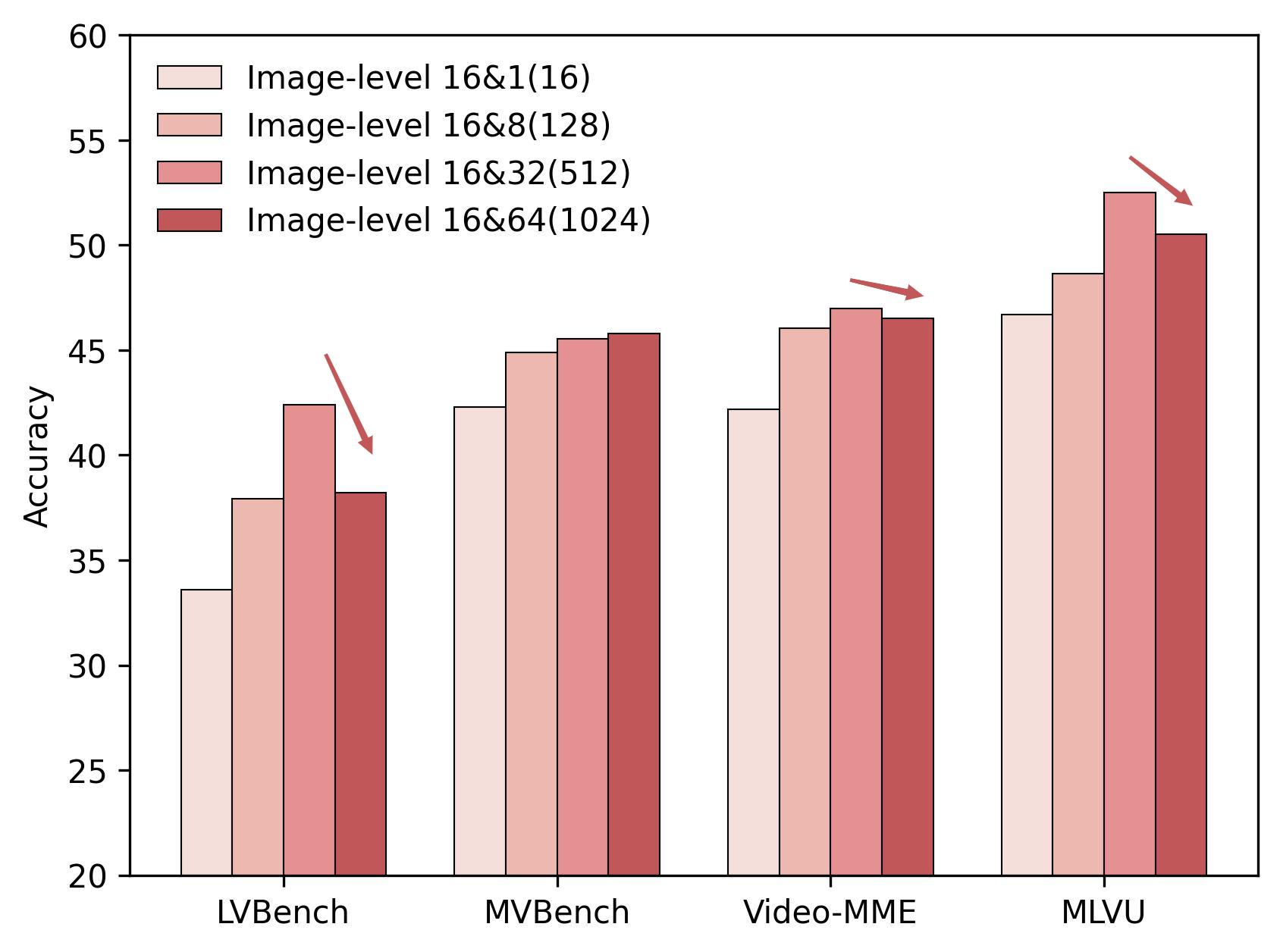}
   \caption{The redundancy phenomenon in image-level resampler shows that as the number of queries increases, the model's performance does not increase indefinitely. '16\&8(128)' represents sampling 16 frames, with each image resampled to 8 queries, and a total of 128 queries.}
   \label{fig:increase}
   \vspace{-0.2in}
\end{wrapfigure}
This approach extracts global features of the sequence while preserving temporal information. Its simple structure ensures that the total number of visual tokens passed to the language model is controlled and limited, regardless of the number of sampled video frames.

\begin{figure}[t]
  \centering
  \begin{subfigure}[b]{0.32\textwidth}
    \includegraphics[width=\linewidth]{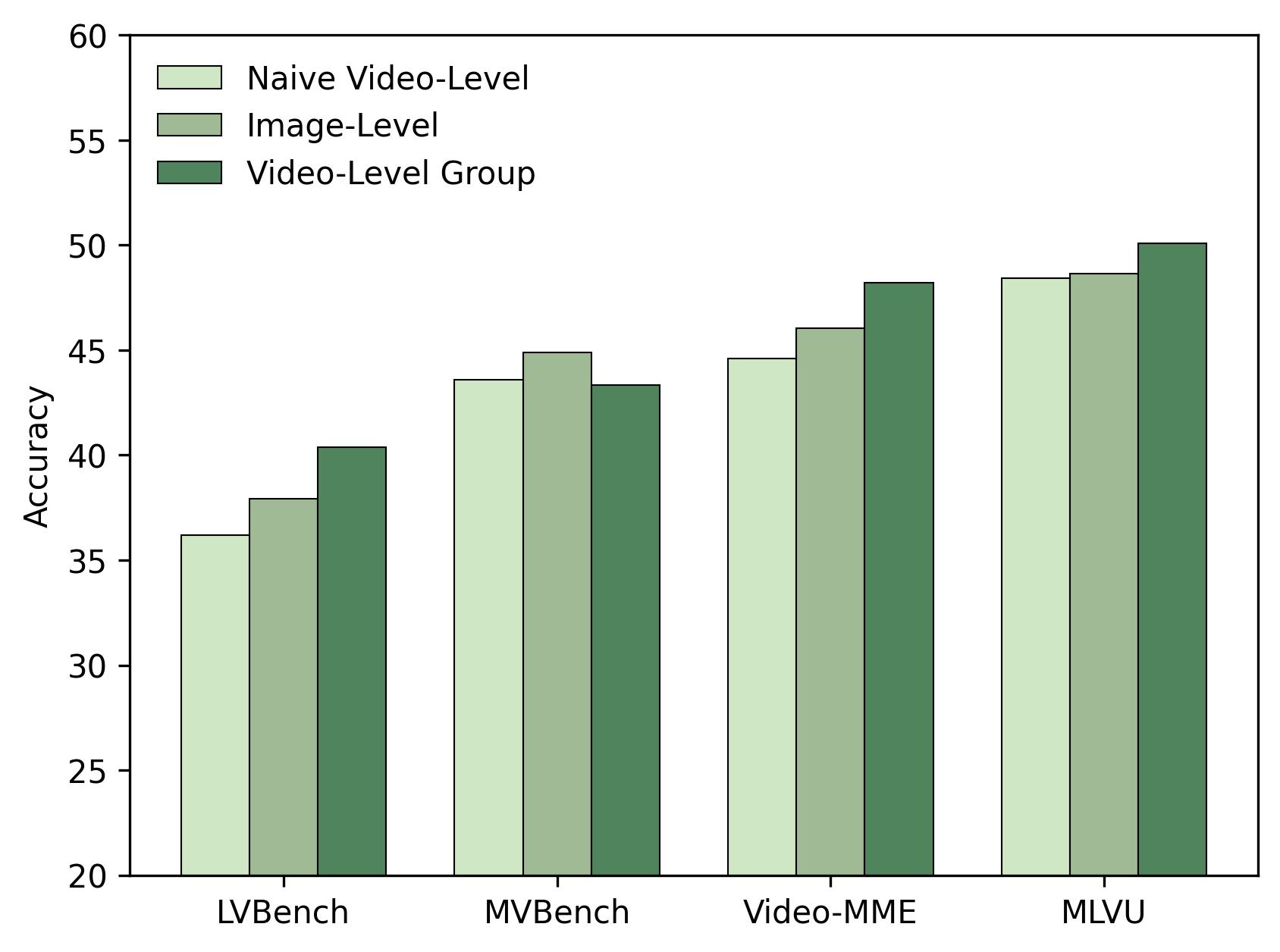}
    \caption{\small Total number of queries = 128}
    \label{fig:box128}
  \end{subfigure}
  \hfill
  \begin{subfigure}[b]{0.32\textwidth}
    \includegraphics[width=\linewidth]{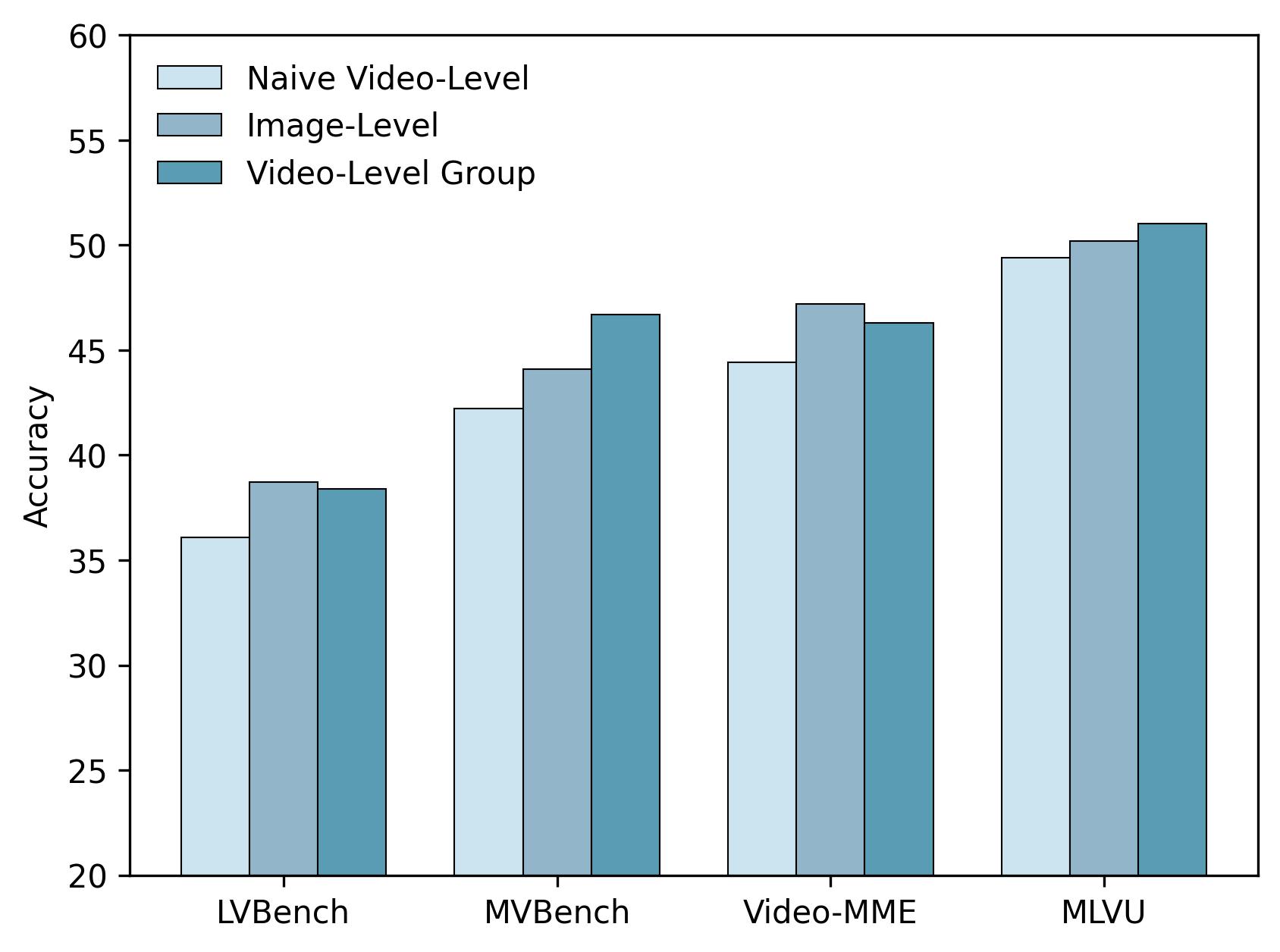}
    \caption{\small Total number of queries = 256}
    \label{fig:box256}
  \end{subfigure}
  \hfill
  \begin{subfigure}[b]{0.32\textwidth}
    \includegraphics[width=\linewidth]{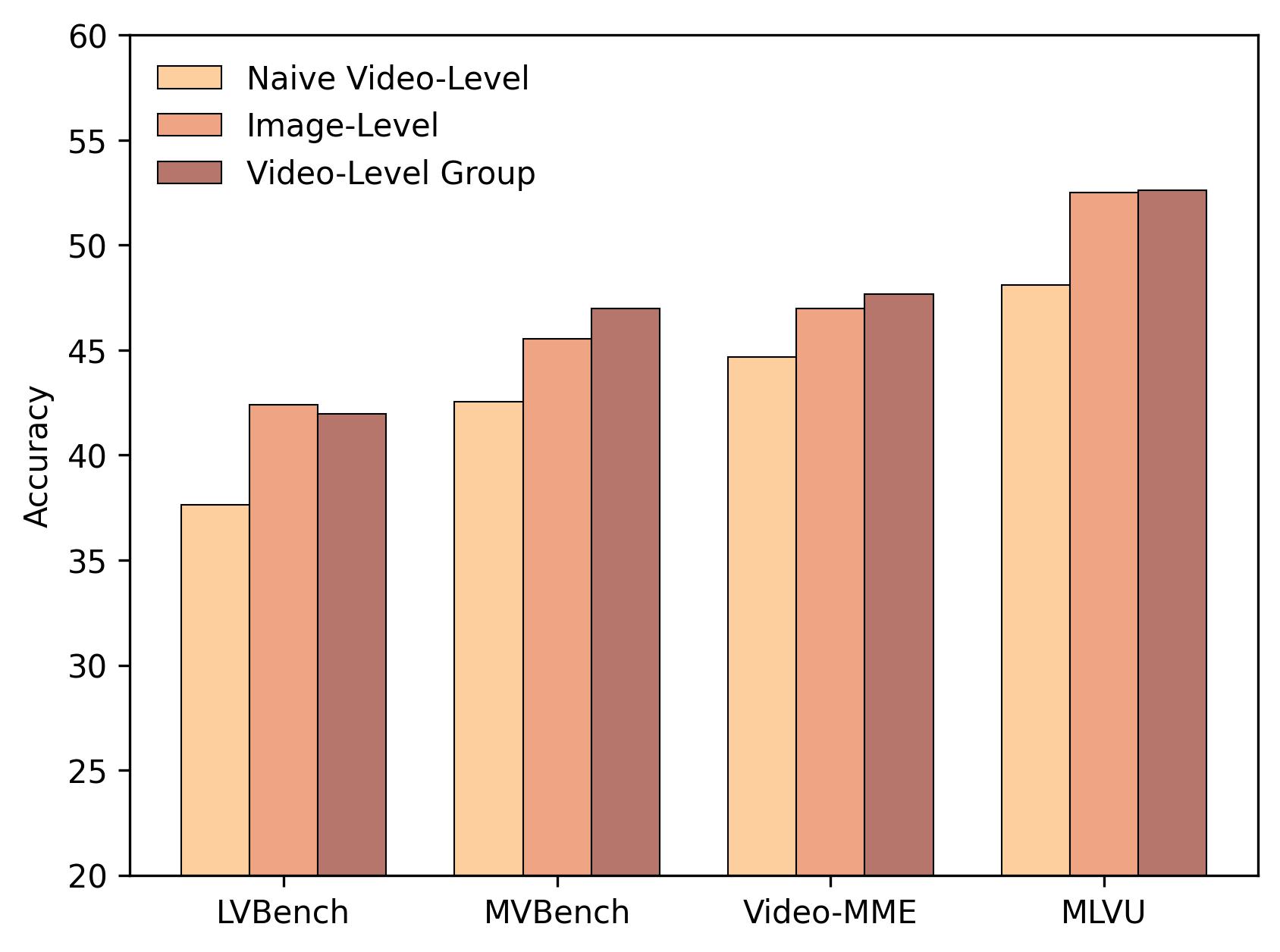}
    \caption{\small Total number of queries = 512}
    \label{fig:box512}
  \end{subfigure}
  \caption{Performance comparison of three resampler with the same total number of queries. Video-level group method performs the best on most benchmarks.}
  \label{fig:box}
  \vspace{-0.2in}
\end{figure}

We first conduct experiments with these two methods, their performance comparison on multiple benchmarks is shown in Fig.~\ref{fig:box}. We find that although the naive video-level resampler is simpler and requires less training time, the image-level resampler consistently outperforms the naive video-level resampler when the total number of learnable queries is the same.

To explore why the image-level resampler performs better when the total number of learnable queries is the same, we also visualize the attention map between the learnable queries and the visual token sequence in the last layer of the naive video-level resampler, as shown in Fig.~\ref{fig:heatmap_naive}. The learnable queries with higher attention are evenly distributed throughout the sequence. This indicates that all the learnable queries pay adequate attention to the entire input sequence, which may lead to redundancy as different queries acquire similar information~\cite{bian2021attention}.

To further validate this hypothesis, we perform varying degrees of zeroing out in the learnable queries in both naive video-level resampler and image-level resampler to investigate their impact on model performance, as illustrated in Fig.~\ref{fig:line_three}. For the image-level resampler, as the proportion of active learnable queries decreased, the model performance also gradually declined. In contrast, for the naive video-level resampler, even retaining only 25\% of the learnable queries maintained the model performance at around 95\%. It was only when further reducing the active learnable queries that the model performance dropped significantly.

\begin{wrapfigure}{r}{0.4\textwidth}
  \vspace{-0.2in}
  \centering
  \includegraphics[width=\linewidth]{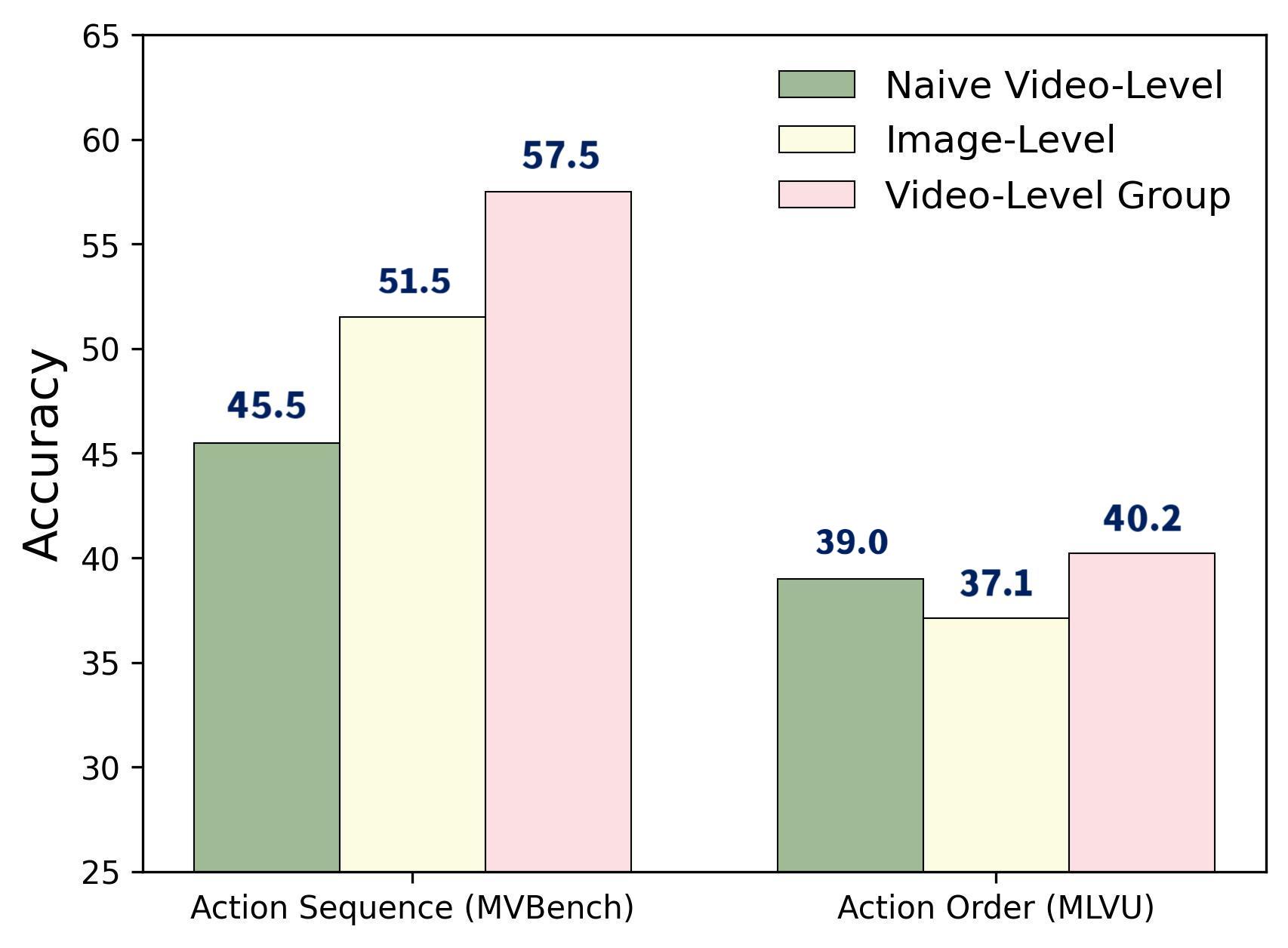}
   \caption{Video-level group resampler demonstrates significantly better performance on temporal comprehension tasks.}
   \label{fig:action}
   \vspace{-0.1in}
\end{wrapfigure}

Based on above observations, we believe that in the naive video-level resampler, the learnable queries must attend to the entire visual token sequence. This approach leads to learnable queries focusing on the global context, which may result in redundancy as different queries acquire similar information. In contrast, the image-level resampler allows the queries to focus specifically on the information relevant to each individual image. However, as the number of queries assigned to each image gradually increases, the image-level method also suffers from redundancy, and the model performance even degrades, as shown in Fig.~\ref{fig:increase}.

Therefore, we aim to develop a resampler that maintains controllable and limited numbers of visual tokens, while mitigating the impact of the number of sampled frames on the total token count. Meanwhile, this method should avoid attention redundancy by enabling the learnable queries to form more effective combinations and focus on the temporal information across video frames, thereby enhancing model performance. To achieve this, we propose the video-level group resampler as follows.

\vspace{-0.05in}
\paragraph{Video-Level Group Resample.} In the video-level group resampler, the visual sequence is first concatenated in the same manner as in the naive video-level resampler before being fed into the connector: 
\begin{equation} 
\hat{H} = \mathit{Linear}(\mathit{Concat}({{H}_1, {H}_2, \dots, {H}_N })) . \label{eq:6} 
\end{equation} 
which ensures that all visual tokens are initially combined into a unified sequence, allowing the model to process the entire visual input. Next, we evenly divide the long sequence and learnable queries into M groups, and each group of queries then learns the corresponding visual features through cross-attention:
\begin{equation}
{\hat{H}_1, \hat{H}_2, \dots, \hat{H}_M } = \mathit{Split}(\hat{H}) . \label{eq:7}
\end{equation}
\begin{equation}
{{Q}_1, {Q}_2, \dots, {Q}_M } = \mathit{Split}(Q) . \label{eq:8} 
\end{equation}
\begin{equation}
\hat{Q}_i = \mathit{CrossAttention}(\{q={Q}_i,\,k,v=\hat{H}_i \}) .
\label{eq:9}
\end{equation}
These operations aim to avoid the problem of attention redundancy. By splitting the sequence and queries, we ensure that each group of queries can focus more effectively on a subset of visual information, which prevents excessive overlap in the attention and allows for more efficient learning. Finally, the outputs of all groups of learnable queries are concatenated as $\hat{Q} = \mathit{Concat}({\hat{Q}_1, \hat{Q}_2, \dots, \hat{Q}_M })$ to maintain the same number.


We also visualize the attention between the learnable queries and the visual token sequence under video-level group resampler as shown in Fig.~\ref{fig:heatmap_group}, and performed varying degrees of zeroing out in the queries to observe the decline in model performance as shown in Fig.~\ref{fig:line_three}. The high-attention values for each group of queries are distributed only across the corresponding visual token sequence, and as the number of normal queries decreases, the model performance gradually declines. These results indicate that the queries with this method have high learning efficiency and do not exhibit significant redundancy. Meanwhile, as illustrated in the case in Fig.~\ref{fig:attncase}, this method is also more sensitive to the temporal relationships between the video frames.

\begin{wrapfigure}{r}{0.45\textwidth}
\vspace{-0.2in}
  \centering
  \includegraphics[width=\linewidth]{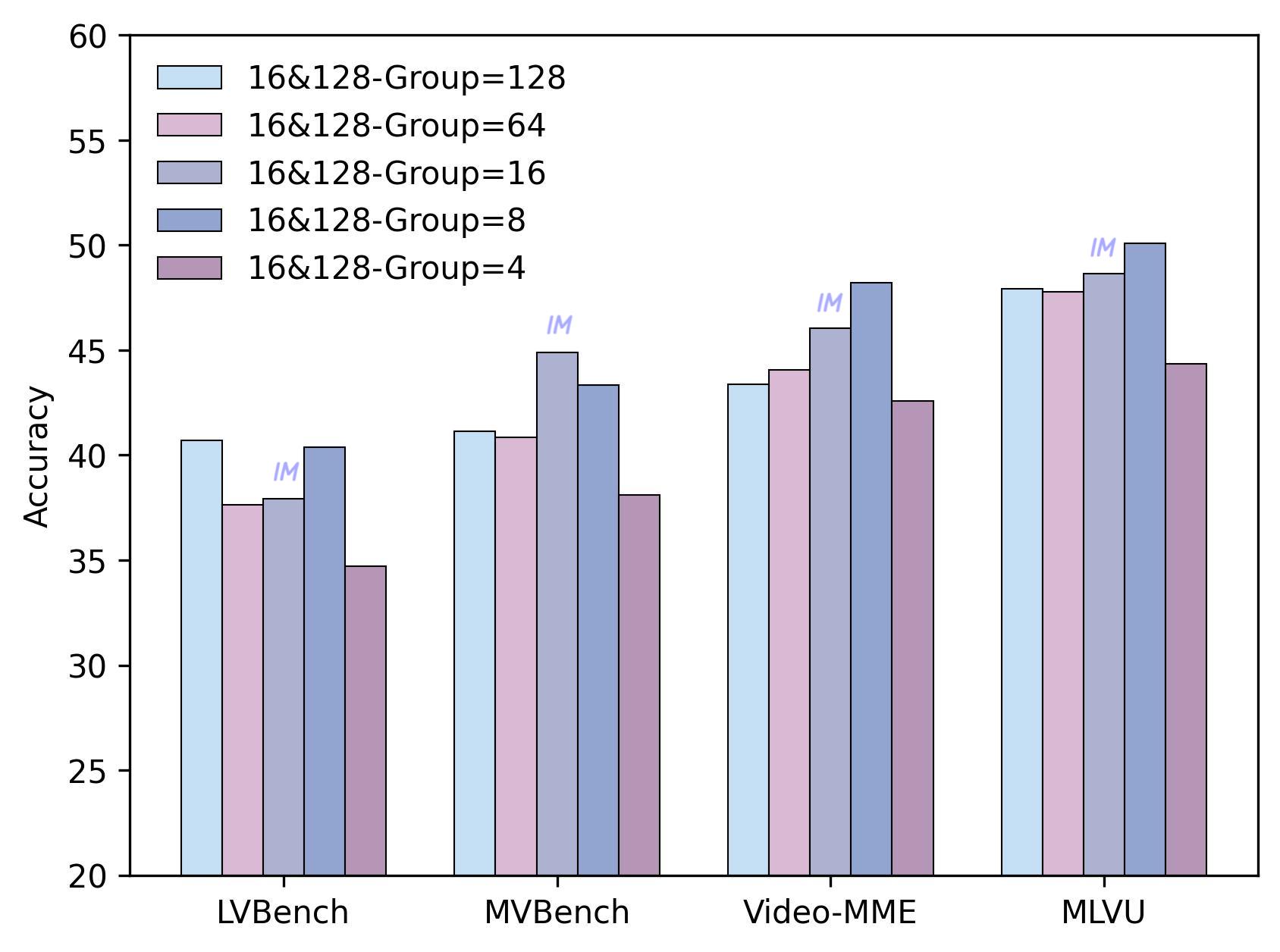}
   \caption{The impact of different group settings on model performance. '16\&128' represents uniformly sampling 16 frames, with a total of 128 queries, and the "Im" label indicates that this setting is equivalent to image-level method.}
   \label{fig:box_group}
   \vspace{-0.15in}
\end{wrapfigure}

By using the video-level group resampler, we keep the total number of visual tokens controllable without increasing as more frames are sampled. This approach also enables the learnable queries to focus on relevant visual information, avoids redundancy, and encourages the model to better capture the temporal relationships between video frames.

We conduct experiments using different resamplers under the same settings, and the results are shown in Fig.~\ref{fig:box}. By grouping learnable queries to improve their learning efficiency, the video-level group method further enhances the model's performance. We also compare the temporal comprehension capabilities of different resampling methods and visualize their attention, as shown in Fig.~\ref{fig:attncase}. Due to the lower learning efficiency of queries in naive video-level method, their attention is sparser, and their ability to capture image information is weaker. In image-level method, the queries focus only on the corresponding image information, leading to weaker temporal comprehension and an inability to correctly interpret the action sequences of characters in the video. In comparison, video-level group method not only has a stronger ability to capture image information, but also better understands the temporal information in the video. As shown in Fig.~\ref{fig:action}, it exhibits a clear advantage in tasks that require greater attention to temporal relationships. 

\begin{wrapfigure}{r}{0.45\textwidth}
\vspace{-0.15in}
  \centering
  \includegraphics[width=\linewidth]{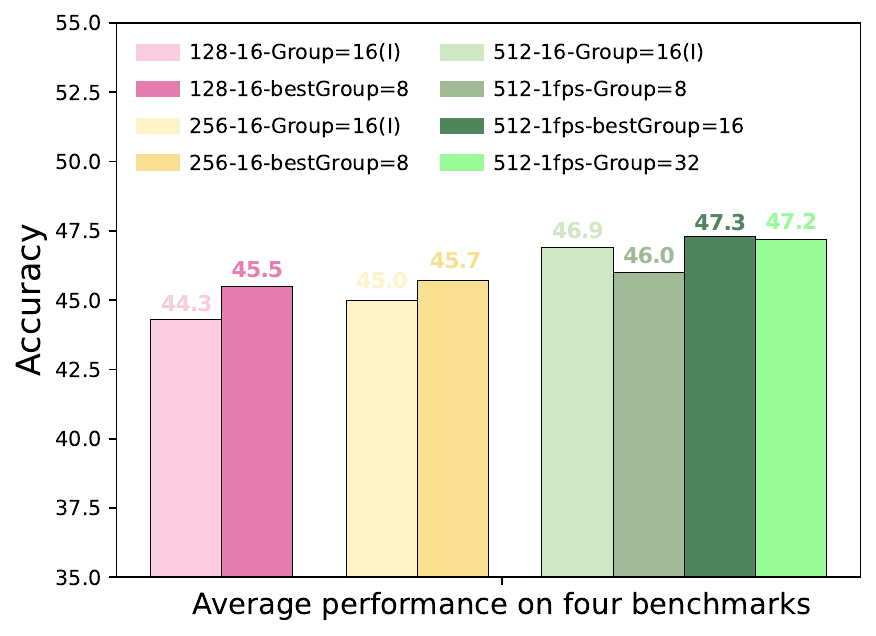}
   \caption{The average performance of the model across four benchmarks under different grouping settings. "128-16-Group=16(I)" indicates a total of 128 queries with 16 uniformly sampled frames and 16 groups, which is equivalent to the Image-Level Method.}
   \label{fig:average}
   \vspace{-0.1in}
\end{wrapfigure}

We also explore the optimal number of groups in the video-level group method, with the results shown in Fig.~\ref{fig:box_group}. In order to explore the upper bound of query-learning efficiency, we experiment with a large number of groups, even setting each group to contain only one query responsible for learning a portion of the visual information. However, the results show that having too few queries per group does not improve the model performance. The best results are achieved when each group contains 16 to 32 queries. 

We then gradually reduce the number of groups, and the model achieves the best performance when each group of queries is responsible for learning information from approximately 2 to 4 frames, as shown in Fig.~\ref{fig:average}. Under the same total number of queries, it demonstrates superior performance and stronger temporal comprehension compared to the setting where each group of queries is responsible for only one image, i.e. the image-level method. The detailed experimental results are presented in Table~\ref{tab:setgroup} of the Appendix.

\vspace{-0.1in}
\subsection{Training Pipeline and Data}

\subsubsection{Pipeline}
\vspace{-0.05in}
The training pipeline of TinyLLaVA-Video is the same as TinyLLaVA~\cite{zhou2024tinyllava}, consisting of two stages:
\vspace{-0.1in}
\paragraph{Stage 1: Pre-training for Feature Alignment.} We keep the visual encoder and language model frozen, training only the connector (resampler) \( P_\phi \) to align the visual embedding \(\hat{Q}\) with the text embedding by video-caption data. 
\vspace{-0.1in}
\paragraph{Stage 2: Supervised Fine-tuning.} We only keep the visual encoder frozen, use the video-text pair (X, Y) in multi-turn conversation, fully fine-tuning small-scale LLM \( F_\theta \) and resampler \( P_\phi \). 

\vspace{-0.05in}
\subsubsection{Data Composition}
\vspace{-0.05in}
The training data for TinyLLaVA-Video is a combination of two datasets: LLaVA-Video-178K~\cite{zhang2024video} and Valley~\cite{luo2023valley}. We selected portions of data from each dataset, with the data sources and quantities used during the pre-training and fine-tuning stages shown in the Table~\ref{tab:traindata}. 

\begin{wraptable}{r}{0.6\textwidth}
\vspace{-0.1in}
\caption{Datasets used for training TinyLLaVA-Video}
\centering
\begin{tabular}{ccc}
\toprule
Stage & Source & \#Sample \\
\arrayrulecolor{gray}
\midrule
Pre-training & LLaVA-Video-178K + Valley & 397K \\
Fine-tuning & LLaVA-Video-178K & 491K \\
\bottomrule
\end{tabular}
\label{tab:traindata}
\vspace{-0.1in}
\end{wraptable}

LLaVA-Video-178K contains multiple subsets, with video captions that are highly detailed and rich in content. The length of the captions is positively correlated with the duration of the videos. Due to the token limitations of small-scale language models and to maintain manageable training time under limited computational resources, we choose videos with durations of no more than 1 minute from the "academic" and "youtube" subsets as training data. Valley is sourced from publicly available video websites. It contains various scene-descriptive video-caption data, and all the videos we use are within 5 minutes in duration. 

\vspace{-0.05in}
\subsubsection{Data Curation}
\label{DataCuration}
\vspace{-0.05in}
For LLaVA-Video-178K, we only remove the corrupted videos, resulting in 108k videos. Since each video corresponds to one video-caption pair and a set of multi-turn dialogue data, the dataset includes 108k video-caption pairs for pre-training and 491k dialogue questions for fine-tuning.

For Valley, the original dataset consisted of approximately 1,406k videos. To simplify and refine the dataset, we perform the following steps: first, we remove corrupted videos or those with incomplete captions. Furthermore, we observe that some captions included descriptions of filming time, angles, and geographic locations. We consider such data excessive and unhelpful for describing video events, so we filter them out as well, ultimately obtaining 289k video-caption data for pre-training. We also compare the impact of using the unfiltered Valley dataset versus the processed dataset as pre-training data on model performance, with the results presented in Table~\ref{tab:dataqua} of the Appendix.
\begin{table}[t]
\caption{The performances on multiple benchmark. TinyLLaVA-Video significantly outperforms these open-source 7B+ models with less training data. The best results are indicated by \textbf{boldface}.}
\vspace{2mm}
\centering
\setlength{\tabcolsep}{5.5pt}
\renewcommand{\arraystretch}{1.2}
\begin{tabular}{lccccccc}
    \toprule
    Model & LLM size & \#Frame & Train Data & MVBench & VM & MLVU & LVB \\
    \midrule
    Video-ChatGPT~\cite{maaz2023video} & 7B & 100 & 1.3M & 32.7 & - & 31.3 & - \\  
    LLaMA-VID~\cite{li2025llama} & 7B & 1fps & 1.5M & 41.4 & - & 33.2 & - \\
    VideoChat2~\cite{li2023videochat} & 7B & 16 & 1.9M & 44.5 & 39.5 & - & 39.3 \\   
    VideoLLaVA~\cite{lin2023video} & 7B & 8 & 2.0M & - & 39.9 & 47.3 & 39.1 \\
    ShareGPT4Video~\cite{chen2024sharegpt4video} & 8B & 16 & 4.8M & - & 39.9 & 46.4 & 39.7 \\
    LLaVA-Mini~\cite{zhang2025llava} & 7B & 1fps & 1.2M & 44.5 & - & 42.8 & - \\
    \midrule
    \textbf{TinyLLaVA-Video} & 3B & 16 & 888K & 45.5 & 47.0 & 52.5 & \textbf{42.4} \\
    \textbf{TinyLLaVA-Video} & 3B & 1fps & 888K & \textbf{47.0} & \textbf{47.7} & \textbf{52.6} & 42.0 \\
    \bottomrule
\end{tabular}
\vspace{-0.1in}
\label{tab:performance}
\end{table}

\vspace{-0.1in}
\section{Experiments}

\vspace{-0.05in}
\subsection{Experimental Setting}
\label{Setting}

\vspace{-0.05in}
\paragraph{Model Architectures.} We first explore the optimal choice of model components. Under the same settings, we conduct experiments with multiple small-scale language models, including Phi-2\cite{javaheripi2023phi} and Qwen2.5-3B~\cite{hui2024qwen2}. The experimental results demonstrate the superiority of Qwen2.5-3B among these language models. Similarly, we experiment with SigLIP~\cite{zhai2023sigmoid}, CLIP~\cite{radford2021learning} and Dinov2~\cite{oquab2023dinov2} as vision encoders, SigLIP achieving the best performance. The results of the experiment are presented in Tables\ref{tab:llm} and~\ref{tab:vt} of the Appendix. Therefore, in subsequent experiments, we will maintain consistency in model components, using Qwen2.5-3B as the language model and SigLIP as the vision encoder.

\vspace{-0.1in}
\paragraph{Training and Evaluation.} For training, TinyLLaVA-Video adopts a two-stage training paradigm, with specific experimental settings presented in Table~\ref{tab:setting} of the Appendix. For evaluation, we selected four widely used benchmarks: Video-MME~\cite{fu2024video}, LongVideoBench~\cite{wu2024longvideobench}, MVBench~\cite{li2024mvbench} and MLVU~\cite{zhou2024mlvu}. These benchmarks cover video lengths ranging from a few seconds to several hours, providing a comprehensive assessment of the model's ability to understand short and long videos. 

\begin{figure}[t]
  \centering
  \includegraphics[width=\textwidth]{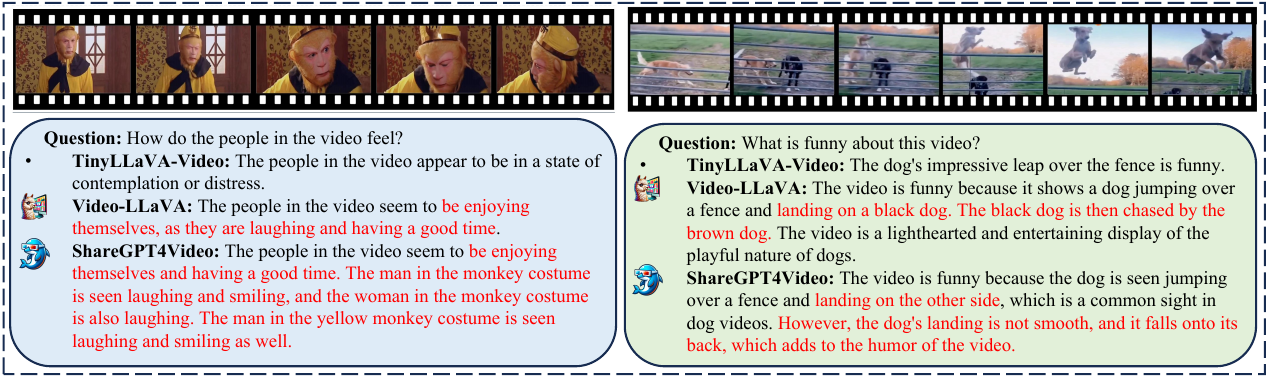}
   \caption{The case demonstrates the understanding capability of TinyLLaVA-Video and compares it with existing video understanding models. The content marked in red indicates understanding errors.}
   \label{fig:case}
   \vspace{-0.15in}
\end{figure}

\vspace{-0.05in}
\subsection{Comparison with Other Open-Source Video-based Models}
\label{Comparison}
\vspace{-0.05in}

\begin{wraptable}{r}{0.42\textwidth}
\vspace{-0.15in}
\caption{Comparison of the efficiency of three resampling methods. The settings for all methods are the same: the total number of queries is 128, and each method samples 16 frames. The result is measured on 8 A100-40G GPUs.}
\centering
\setlength{\tabcolsep}{2.5pt}
\renewcommand{\arraystretch}{0.7}
\begin{tabular}{ccc}
    \toprule
    Resampler & Train Time & VRAM \\
    \arrayrulecolor{gray}
    \toprule
    Naive Video-Level & 20H & 15.3GB \\   
    Image-Level & 25H & 14.9GB  \\   
    Video-Level Group & 25H & 14.9GB  \\   
    \bottomrule
\end{tabular}
\vspace{-0.05in}
\label{tab:effi}
\vspace{-0.1in}
\end{wraptable}

Finally, we compare TinyLLaVA-Video with recent fully open-source advanced models on multiple benchmarks. As shown in Table~\ref{tab:performance}, despite having fewer parameters and using comparable or even smaller-scale training data than these 7B models, our model still achieves better performance on both comprehensive video understanding and long video benchmarks. In addition to quantitative evaluations, Fig.~\ref{fig:case} presents cases in which our model effectively captures implicit information in videos, such as character emotions, and accurately understands the interesting aspects of the video. Detailed experimental results and more cases are presented in Table~\ref{tab:lvbench} to~\ref{tab:mlvu} and Fig.~\ref{fig:casetwo} of the Appendix. Therefore, we believe that the capabilities of TinyLLaVA-Video make it a strong platform for training and researching video models for researchers with limited resources.

\vspace{-0.05in}
\subsection{Efficiency}
\vspace{-0.05in}

We also compare the training time and inference memory usage of the three resampler. As shown in Table~\ref{tab:effi}, the image-level method and video-level group method are comparable. Without quantization, TinyLLaVA-Video can inference on NVIDIA 3090 GPUs or even NVIDIA 4060 Ti GPUs.

\vspace{-0.05in}
\section{Limitations}
\vspace{-0.05in}
To maintain manageable training time with limited computational resources, we did not use a large amount of data when training TinyLLaVA-Video, only fine-tuned it by filtered multi-turn dialogues. However, since many current benchmarks are in the form of multiple-choice questions, we believe that incorporating data of similar formats during fine-tuning can further improve the model’s performance. It will be part of our future work.

\section{Conclusion}
\vspace{-0.05in}
We propose TinyLLaVA-Video, a smaller LMM that processes video sequences with video-level group resampler. This novel approach significantly reduces the number of visual tokens input to the language model, effectively avoiding redundancy and demonstrating superior capabilities in information capture and temporal comprehension. The model can be trained in just one day on 8 A100-40G GPUs. Under limited computational resources, our best model outperforms several existing 7B models across multiple benchmarks. The potential of small-scale video understanding LMMs still has significant room for exploration, and we hope our work can serve as a foundational reference for practitioners exploring this direction. 


\paragraph{Acknowledgment.} 
This work was partially supported by the National Science and Technology Major Project (Grant No. 2022ZD0116310), National Natural Science Foundation of China (Grant No. 62476016), the Fundamental Research Funds for the Central Universities.
{
    \small
    \bibliographystyle{plain}
    \bibliography{main}
}
\newpage
\appendix

\section{Technical Appendices and Supplementary Material}

\subsection{Detailed Experimental Settings}

\begin{wraptable}{r}{0.58\textwidth}
\vspace{-0.15in}
\caption{Training details of TinyLLaVA-Video.}
\centering
\setlength{\tabcolsep}{4pt}
\begin{tabular}{ccc}
    \toprule
    Setting & Pre-training & Fine-tuning \\
    \arrayrulecolor{gray}
    \toprule
    Batchsize & 128 & 64  \\   
    Learning rate & $1\mathrm{e}{-4}$ & $2\mathrm{e}{-5}$  \\   
    Schedule & \multicolumn{2}{c}{Cosine}  \\   
    Warmup Ratio & \multicolumn{2}{c}{0.03} \\   
    Epoch & \multicolumn{2}{c}{1} \\
    Optimization strategy & \multicolumn{2}{c}{deepseed ZeRO-3} \\
    \bottomrule
\end{tabular}
\label{tab:setting}
\vspace{-0.1in}
\end{wraptable}

As mentioned in section~\ref{Setting}, our model follows the two-stage training paradigm similar to TinyLLaVA~\cite{zhou2024tinyllava}. During pre-training, we update only the connector \( P_\phi \) while keeping the rest of the model frozen, maintaining a learning rate of $1\mathrm{e}{-4}$ and a batch size of 128. In the fine-tuning stage, we freeze the vision encoder \( V_\varphi \) and update both the connector \( P_\phi \) and the small-scale LLM \( F_\theta \) using a learning rate of $2\mathrm{e}{-5}$ and a batch size of 64. All experimental results presented adhere to the settings shown in Table~\ref{tab:setting}.

\begin{wraptable}{r}{0.6\textwidth}
\vspace{-0.15in}
\caption{The performances on LongVideoBench validation.}
\centering
\setlength{\tabcolsep}{5pt}
\renewcommand{\arraystretch}{0.9}
\begin{tabular}{cccc}
    \toprule
    Model & LLM size & \#Frame & Avg \\
    \arrayrulecolor{gray}
    \toprule
    VideoLLaVA~\cite{lin2023video} & 7B & 8 & 39.1  \\   
    mPLUG-Owl2~\cite{ye2024mplug} & 7B & 8 & 39.1  \\   
    VideoChat2-Mistral~\cite{li2023videochat} & 7B & 16 & 39.3  \\   
    ShareGPT4Video~\cite{chen2024sharegpt4video} & 8B & 16 & 39.7 \\   
    PLLaVA~\cite{xu2024pllava} & 7B & 32 & 40.2 \\
    LLaVA-1.5~\cite{liu2024visual} & 7B & 8 & 40.3 \\
    \arrayrulecolor{gray}
    \midrule
    \textbf{TinyLLaVA-Video} & 3B & 16 & \textbf{42.4} \\
    \textbf{TinyLLaVA-Video} & 3B & 1fps & 42.0 \\
    \bottomrule
\end{tabular}
\vspace{-0.1in}
\label{tab:lvbench}
\end{wraptable}

Additionally, for fps sampling in the experiment, since the maximum duration of the training data is only five minutes, we normally sample one frame per second during training. However, during testing, some long videos can last several hours, and sampling one frame per second would significantly increase the testing time. Therefore, we limit the maximum number of sampled frames to 64 during testing.

\subsection{Detailed Experimental Results}

\begin{table*}[htbp]
\caption{The performances on MVBench. "Attri", "Chara" and "Cog" represent Attribute, Character and Cognition, respectively. The best results are indicated by \textbf{boldface}.}
\centering
\footnotesize
\setlength{\tabcolsep}{1.8pt}
\renewcommand{\arraystretch}{1}
\begin{tabular}{ccccccccccccc}
    \toprule
    Model & LLM size & \#Frame & Action & Object & Position & Scene & Count & Attri & Pose & Chara & Cog & Avg \\
    \arrayrulecolor{gray}
    \midrule
    Otter-I~\cite{li2023otter} & 7B & 16 & 35.5 & 40.7 & 22.3 & 55.0 & 26.3 & 28.5 & 22.3 & 27.0 & 32.5 & 33.5 \\  
    mPLUG-Owl-V~\cite{ye2024mplug} & 7B & 16 & 28.4 & 33.0 & 25.0 & 29.0 & 29.3 & 40.0 & 34.0 & 31.0 & 25.3 & 29.7 \\   
    Video-ChatGPT~\cite{maaz2023video} & 7B & 100 & 32.1 & 40.7 & 21.5 & 31.0 & 28.0 & 39.5 & 38.8 & 33.0 & 30.3 & 32.7 \\   
    Video-LLaMA~\cite{zhang2023video} & 7B & 16 & 34.4 & 42.2 & 22.5 & 43.0 & 28.3 & 32.5 & 39.0 & 40.0 & 29.3 & 34.1 \\   
    VideoChat~\cite{li2023videochat} & 7B & 16 & 38.0 & 41.2 & 26.3 & 48.5 & 27.8 & 42.5 & 26.3 & 41.0 & 27.7 & 35.5 \\   
    LLaMA-VID~\cite{li2025llama} & 7B & 1fps & 48.1 & 42.4 & 22.8 & 84.5 & 35.3 & 44.5 & 31.3 & 39.0 & 38.3 & 41.4 \\
    LLaVA-Mini~\cite{zhang2025llava} & 8B & 1fps & 52.1 & 43.2 & \textbf{31.8} & \textbf{85.5} & \textbf{37.5} & 44.5 & 29.5 & \textbf{52.0} & 35.0 & 44.5 \\
    \arrayrulecolor{gray}
    \midrule
    \textbf{TinyLLaVA-Video} & 3B & 16 & 55.1 & 48.8 & 30.5 & 81.5 & 30.8 & \textbf{48.0} & \textbf{41.3} & 38.5 & 38.5 & 45.5 \\
    \textbf{TinyLLaVA-Video} & 3B & 1fps & \textbf{58.2} & \textbf{51.0} & 30.8 & \textbf{85.5} & 33.8 & 45.5 & 40.0 & 39.5 & \textbf{38.7} & \textbf{47.0} \\
    \bottomrule
\end{tabular}
\label{tab:mvbench}
\vspace{-0.1in}
\end{table*}

\begin{table*}[htbp]
\caption{The performances on Video-MME (without subtitle). TinyLLaVA-Video significantly outperforms these 7B+ models. The best results are indicated by \textbf{boldface}.}
\centering
\footnotesize
\setlength{\tabcolsep}{2pt}
\renewcommand{\arraystretch}{1}
\begin{tabular}{ccccccc}
    \toprule
    Model & LLM size & \#Frame & Short (\textless2min) & Medium (4$\sim$15min) & Long (30$\sim$60min) & Avg \\
    \arrayrulecolor{gray}
    \midrule
    VideoLLaVA~\cite{lin2023video} & 7B & 8 & 45.3 & 38.0 & 36.2 & 39.9  \\   
    Qwen-VL-Chat~\cite{bai2023qwen} & 7B & 4 & 46.9 & 38.7 & 37.8 & 41.1  \\   
    ST-LLM~\cite{liu2025st} & 7B & 64 & 45.7 & 36.8 & 31.3 & 37.9  \\   
    ShareGPT4Video~\cite{chen2024sharegpt4video} & 8B & 16 & 48.3 & 36.3 & 35.0 & 39.9  \\   
    VideoChat2-Mistral~\cite{li2023videochat} & 7B & 16 & 48.3 & 37.0 & 33.2 & 39.5  \\ 
    Chat-UniVi-v1.5~\cite{jin2024chat} & 7B & 64 & 45.7 & 40.3 & 35.8 & 40.6  \\  
    SliME~\cite{zhang2024beyond} & 8B & 8 & 53.3 & 42.7 & 39.8 & 45.3  \\
    \arrayrulecolor{gray}
    \midrule
    \textbf{TinyLLaVA-Video}& 3B & 16 & 55.9 & \textbf{46.6} & 38.4 & 47.0 \\
    \textbf{TinyLLaVA-Video}
    & 3B & 1fps & \textbf{57.1} & 46.0 & \textbf{39.9} & \textbf{47.7} \\
    \bottomrule
\end{tabular}
\label{tab:videomme}
\vspace{-0.1in}
\end{table*}

\begin{table*}[t]
\caption{The performances on MLVU (Dev). Evaluation includes holistic LVU tasks (TR: Topic Reasoning, AR: Anomaly Recognition), single-detail LVU tasks (NQA: NeedleQA, ER: Ego Reasoning, PQA: PlotQA) and multi-detail LVU tasks (AO: Action Order, AC: Action Count). The best results are indicated by \textbf{boldface}.}
\footnotesize
\centering
\setlength{\tabcolsep}{5pt}
\renewcommand{\arraystretch}{1}
\begin{tabular}{ccccccccccc}
    \toprule
    \multirow{2}{*}{Model} & \multirow{2}{*}{LLM Size} & \multirow{2}{*}{\#Frame} & \multicolumn{2}{c}{Holistic} & \multicolumn{3}{c}{Single Detail} & \multicolumn{2}{c}{Multi Detail} & \multirow{2}{*}{Avg} \\
    \cmidrule(lr){4-5} \cmidrule(lr){6-8} \cmidrule(lr){9-10} & & & TR & AR & NQA & ER & PQA & AO & AC & \\
    \arrayrulecolor{gray}
    \midrule
    LLaVA-1.6~\cite{liu2024llava} & 7B & 16 & 60.6 & 41.0 & 43.1 & 38.4 & 41.0 & 25.5 & 25.7 & 39.3 \\ 
    LLaMA-VID~\cite{li2025llama} & 7B & 1fps & 50.8 & 34.5 & 30.1 & 32.7 & 32.5 & 23.9 & 27.8 & 33.2  \\   
    Video-LLaMA-2~\cite{cheng2024videollama} & 13B & 16 & 54.5 & 41.5 & 39.4 & 33.5 & 35.4 &  18.5 & 25.7 & 35.5  \\   
    Video-ChatGPT~\cite{maaz2023video} & 7B & 100 & 26.9 & 24.0 & 40.3 & 42.0 & 29.9 & 25.1 & 31.1 & 31.3  \\   
    VideoChat2-Vicuna~\cite{li2023videochat} & 7B & 16 & 74.6 & 51.5 & 42.0 & 47.4 & 43.8 & 22.8 & 29.6 & 44.5  \\   
    MiniGPT4-Video~\cite{ataallah2024minigpt4} & 7B & 90 & 70.9 & 52.5 & 49.0 & 48.6 & 44.5 &  23.2 & 23.0 & 44.5  \\   
    VideoLLaVA~\cite{lin2023video} & 7B & 8 & 71.6 & \textbf{57.0} & 53.2 & 45.2 & 48.4 & 20.1 & \textbf{35.9} & 47.3  \\
    LLaVA-Mini~\cite{zhang2025llava} & 8B & 1fps & 76.0 & 50.0 & 44.5 & 37.5 & 49.0 & 24.3 & 18.4 & 42.8  \\
    \arrayrulecolor{gray}
    \midrule
    \textbf{TinyLLaVA-Video} & 3B & 16 & 74.1 & 56.0 & 63.7 & \textbf{49.7} & \textbf{61.0} & 37.1 & 25.7 & 52.5 \\
    \textbf{TinyLLaVA-Video} & 3B & 1fps & \textbf{77.2} & 51.0 & \textbf{66.8} & 47.7 & 59.4 & \textbf{39.4} & 26.7 & \textbf{52.6} \\
    \bottomrule
\end{tabular}
\vspace{-0.1in}
\label{tab:mlvu}
\end{table*}

\begin{table*}[t]
\caption{More detailed results of video-level group resampler. VM and LVB represent the Video-MME and LongVideoBench, respectively. The best results for each section are indicated by \textbf{boldface}.}
\footnotesize
\centering
\setlength{\tabcolsep}{3pt}
\renewcommand{\arraystretch}{0.9}
\begin{tabular}{cccccccccc}
    \toprule
    \#Frame & \#Total Query & \#Group & \#Query per group & VM (S) & VM (M) & VM (L) & LVB & MVBench & MLVU \\
    \arrayrulecolor{gray}
    \midrule
    \multirow{5}{*}{16} & \multirow{5}{*}{128} & 128 & 1 & 51.7 & 43.8 & 34.7 & 40.7 & 41.1 & 47.9 \\
     &  & 64 & 2 & 51.9 & 44.0 & 36.2 & 37.6 & 40.9 & 47.8 \\
     &  & 16 & 8 & 51.9 & 46.1 & 40.1 & 37.9 & \textbf{44.9} & 48.6 \\
     &  & 8 & 16 & \textbf{55.0} & \textbf{48.9} & \textbf{40.7} & \textbf{40.4} & 43.3 & \textbf{50.1} \\
     &  & 4 & 32 & 50.4 & 42.8 & 34.6 & 34.7 & 38.1 & 44.4 \\
     \arrayrulecolor{gray}
     \midrule
     \multirow{2}{*}{16} & \multirow{2}{*}{256}& 16 & 16 & \textbf{55.6} & 45.1 & \textbf{40.8} & \textbf{38.7} & 44.0 & 50.2 \\
     &  & 8 & 32 & 53.2 & \textbf{45.6} & 40.2 & 38.4 & \textbf{46.9} & \textbf{51.0} \\
     \arrayrulecolor{gray}
     \midrule
     \multirow{3}{*}{16} & \multirow{3}{*}{512} & 32 & 16 & 54.9 & 45.3 & 37.3 & 40.9 & 46.1 & 51.6 \\
      &  & 16 & 32 & \textbf{55.9} & \textbf{46.6} & 38.4 & \textbf{42.4} & \textbf{45.5} & \textbf{52.5} \\
      &  & 8 & 64 & 54.4 & 44.4 & \textbf{38.7} & 36.4 & \textbf{45.5} & 51.0 \\
    \arrayrulecolor{gray}
    \midrule
    \multirow{3}{*}{1fps} & \multirow{3}{*}{512} & 32 & 16 & 56.6 & \textbf{46.2} & 38.6 & \textbf{43.0} & 46.0 & \textbf{52.8} \\
     &  & 16 & 32 & \textbf{57.1} & 46.0 & \textbf{39.9} & 42.0 & \textbf{47.0} & 52.6 \\
     &  & 8 & 64 & 54.4 & 46.1 & 39.1 & 39.0 & 46.2 & 51.5 \\
     \arrayrulecolor{gray}
    \bottomrule
\end{tabular}
\label{tab:setgroup}
\vspace{-0.2in}
\end{table*}

As mentioned in section~\ref{analysis}, we present more detailed experimental results in different grouping settings, as shown in Table~\ref{tab:setgroup}, where proper grouping can lead to better model performance. As mentioned in section~\ref{Comparison}, we compare TinyLLaVA-Video with recent advanced models on multiple benchmarks. Although our model has fewer parameters, as shown in Table~\ref{tab:mvbench} and~\ref{tab:videomme}, it still outperforms several 7B models on comprehensive video understanding benchmarks. Although the training data for TinyLLaVA-Video consists mostly of videos under 1 minute, as shown in Table~\ref{tab:mlvu} and~\ref{tab:lvbench}, our model's long video understanding ability is comparable to several existing 7B models in evaluations on long video benchmarks.

\vspace{-0.05in}
\subsection{Comparison of Different Small-scale LLMs}
\vspace{-0.05in}
\begin{wraptable}{r}{0.65\textwidth}
\vspace{-0.15in}
\caption{Comparison of different small-scale LLMs. The other experimental settings are the same: use SigLIP as vision encoder, uniformly sample 16 frames, and set the number of learnable queries to 512. LVB represents LongVideoBench. The best results are indicated by \textbf{boldface}.}
\centering
\setlength{\tabcolsep}{5pt}
\renewcommand{\arraystretch}{0.8}
\begin{tabular}{ccccc}
    \toprule
    LLM & Video-MME & LVB & MVBench & MLVU \\
    \arrayrulecolor{gray}
    \midrule
    Qwen2.5-3B & \textbf{44.7} & \textbf{37.6} & 42.5 & \textbf{48.1} \\
    Phi2-2.7B & 42.0 & 36.9 & \textbf{43.5} & 46.0 \\
    Qwen2.5-1.5B & 34.5 & 29.5 & 39.0 & 40.5 \\
    Gemma-2B-it & 26.0 & 23.2 & 31.1 & 27.3 \\
    \bottomrule
\end{tabular}
\label{tab:llm}
\vspace{-0.1in}
\end{wraptable}

As mentioned in section~\ref{Setting}, we conducted experiments using naive video-level resampler on different small-scale language models under the same settings. The results are presented in Table~\ref{tab:llm}. It can be observed that when Qwen2.5-3B is used as the language model, the model performs better across multiple benchmarks. Therefore, we choose Qwen2.5-3B as the language model for subsequent experiments.

\subsection{Comparison of Different Vision Encoders}

\begin{wraptable}{r}{0.65\textwidth}
\vspace{-0.15in}
\caption{Comparison of different vision encoders. The other experimental settings are the same: use Qwen2.5-3B as LLM, uniformly sample 16 frames, and set the number of learnable queries to 512. LVB represents LongVideoBench. The best results are indicated by \textbf{boldface}.}
\centering
\setlength{\tabcolsep}{5pt}
\renewcommand{\arraystretch}{0.7}
\begin{tabular}{ccccc}
    \toprule
    Vison Tower & Video-MME & LVB & MVBench & MLVU \\
    \arrayrulecolor{gray}
    \midrule
    SigLIP & \textbf{44.7} & 37.6 & \textbf{42.5} & \textbf{48.1} \\
    CLIP & 38.6 & 35.5 & 39.0 & 46.5 \\
    Dinov2 & 41.3 & \textbf{37.7} & 39.9 & 46.9 \\
    \bottomrule
\end{tabular}
\label{tab:vt}
\vspace{-0.1in}
\end{wraptable}

As mentioned in section~\ref{Setting}, we also conduct experiments with different vision encoders. The results are presented in Table~\ref{tab:vt}. It can be observed that using SigLIP as the vision encoder achieves the best performance, and the model using Dinov2 outperforms the model using CLIP on most benchmarks. Since video understanding still relies on the extraction of image features, using SigLIP with higher input resolution is more helpful for capturing fine-grained visual information. Therefore, we choose SigLIP as the vision encoder for subsequent experiments.

\subsection{Investigating the Effects of Data Quality}

\begin{wraptable}{r}{0.62\textwidth}
\vspace{-0.15in}
\caption{Comparison of different pre-training data. The fine-tuning data and other experimental settings are the same: use Qwen2.5-3B as LLM, SigLIP as vision encoder, uniformly sample 16 frames, and set the number of learnable queries to 512. LVB represents LongVideoBench. The best results are indicated by \textbf{boldface}.}
\centering
\setlength{\tabcolsep}{3.5pt}
\renewcommand{\arraystretch}{1.1}
\begin{tabular}{ccccc}
    \toprule
    Pre-training Data & Video-MME & LVB & MVBench & MLVU \\
    \arrayrulecolor{gray}
    \midrule
    397k Mixed Data & 44.7 & \textbf{37.6} & \textbf{42.5} & \textbf{48.1} \\
    405k Unfil Data & \textbf{44.8} & 36.1 & 40.7 & 47.2 \\
    \bottomrule
\end{tabular}
\label{tab:dataqua}
\vspace{-0.15in}
\end{wraptable}

As mentioned in section~\ref{DataCuration}, we perform pre-training using 397k mixed data and 405k unfiltered Valley data, respectively. The results are presented in Table~\ref{tab:dataqua}. It can be observed that when using data on the same scale, higher quality training data enables the model to achieve better performance on most benchmarks. Therefore, we choose to use 397k mixed data for pre-training in subsequent experiments.

\subsection{More Cases}

\begin{figure*}[h!]
  \centering
  \includegraphics[width=\textwidth]{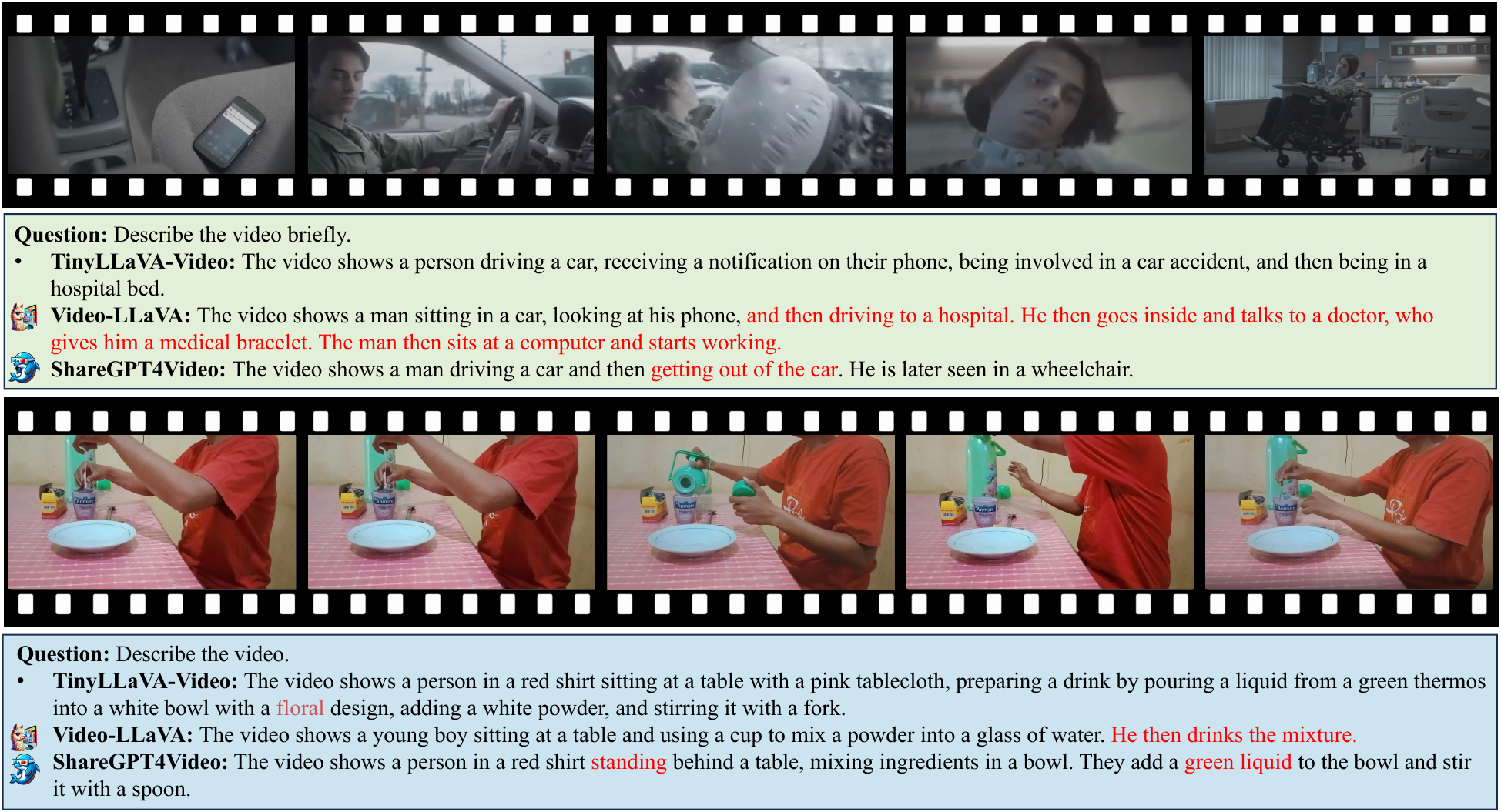}
   \caption{The case demonstrates the understanding capability of TinyLLaVA-Video on different videos and compares it with existing video understanding models. The content marked in red indicates understanding errors.}
   \label{fig:casetwo}
   \vspace{-0.1in}
\end{figure*}

As mentioned in section~\ref{Comparison}, we present more cases. TinyLLaVA-Video is able to accurately capture scene information, as shown in Fig.~\ref{fig:casetwo}. It can correctly interpret the events in the video and describe the sequence of events. In contrast, other models often fail to understand correctly and tend to exhibit severe hallucinations, describing scenes that do not appear in the video. In general, TinyLLaVA-Video demonstrates a stronger video understanding capability.

\subsection{Benchmarks}

We select four benchmarks to evaluate the model, covering both short and long videos, various types of video, and multiple tasks and domains. These four benchmarks provide a comprehensive evaluation of the model's overall capabilities.

\paragraph{Video-MME.} Video-MME is divided into three subsets based on duration, with video lengths ranging from 11 seconds to 1 hour. The dataset includes various types of videos, covering six primary visual domains, including knowledge, film, and sports competitions, as well as 30 subfields. The questions are sufficiently challenging and do not contain QA pairs that can be answered solely based on textual questions.

\paragraph{MVBench.} The video durations in MVBench are mostly less than 1 minute, but the dataset covers 20 challenging video tasks, including action recognition and scene transition. It primarily evaluates the model's understanding of fine-grained information.

\paragraph{MLVU.} The durations of the videos in MLVU range from a few minutes to several hours, with various types of video, including movies, surveillance footage, ego-centric videos and cartoons. It primarily evaluates the model's comprehensive understanding of long videos.

\paragraph{LongVideoBench.} The video durations range from a few seconds to 1 hour, covering 17 fine-grained categories. It evaluates the model's understanding of both short and long videos across multiple domains.
\clearpage

\end{document}